\documentclass[11pt,a4paper]{article}
\usepackage{imubench_arxiv}            
\usepackage[numbers,sort&compress]{natbib} 

\usepackage{amsmath,amsfonts,bm}

\def\eqref#1{equation~\ref{#1}}

\def\1{\bm{1}}

\DeclareMathAlphabet{\mathsfit}{\encodingdefault}{\sfdefault}{m}{sl}
\SetMathAlphabet{\mathsfit}{bold}{\encodingdefault}{\sfdefault}{bx}{n}

\usepackage{needspace}
\usepackage{float}
\usepackage[utf8]{inputenc} 
\usepackage[T1]{fontenc}    
\usepackage{hyperref}       
\usepackage{url}            
\usepackage{booktabs}       
\usepackage{amsmath}
\usepackage{amsfonts}       
\usepackage{amssymb}
\usepackage{nicefrac}       
\usepackage{microtype}      
\usepackage{xcolor}         
\hypersetup{colorlinks=true, linkcolor=linkblue, citecolor=linkblue, urlcolor=linkblue}
\usepackage{graphicx}
\usepackage{placeins}    

\usepackage{subcaption}
\usepackage{wrapfig}      
\usepackage{longtable}
\usepackage{array}
\usepackage{ragged2e}
\usepackage{multirow}
\usepackage{tikz}
\usetikzlibrary{positioning,arrows.meta,calc,fit}

\hypersetup{pdftitle={IMU-Bench: Benchmarking Wrist IMU Representations Across Kinematic, Semantic and Clinical Tasks},
            pdfauthor={Yiquan Li, Aidan Bradshaw, Chenfeng Gao, Karan Ahuja}}
\newcommand{\runninghead}{IMU-Bench: Benchmarking Wrist IMU Representations Across Kinematic, Semantic and Clinical Tasks}

\begin{document}

\vspace*{-25pt}
\begin{tcolorbox}[colback=titleboxbg, colframe=titleboxbg, boxrule=0pt, arc=0pt, sharp corners,
                  left=15pt, right=15pt, top=15pt, bottom=15pt, width=\linewidth]
\begin{flushleft}
{\raggedright\titlefont IMU-Bench: Benchmarking Wrist IMU Representations Across Kinematic, Semantic and Clinical Tasks\par}
\vskip 17pt
{\raggedright\authorfont Yiquan Li\textsuperscript{$*$}, Aidan Bradshaw\textsuperscript{$*$}, Chenfeng Gao, Karan Ahuja\par}
\vskip 0.3em
{\raggedright\affilfont Northwestern University\par}
\end{flushleft}
\vskip 5pt
{\absfont Wrist-worn inertial measurement units (IMUs) provide a common sensing modality for tasks ranging from activity and gesture recognition to pose estimation, inertial odometry and clinical motor assessment. Existing wearable benchmarks evaluate performance across multiple tasks, but leave open how supervision for one task affects transfer to others and whether these tasks benefit from a shared representation. We present IMU-Bench, a benchmark of ten tasks from eight datasets for studying IMU multi-task performance and cross-task transfer. We compare six self-supervised methods pretrained on unlabeled recorded and simulated wrist motion, evaluating classification, pose and velocity estimation, and IMU--language retrieval. We find that our masked autoencoder, InrtlWrist, achieves the best or joint-best frozen-probe performance on nine of ten tasks, and fine-tuning improves on training from scratch on nine of ten tasks, including full-body pose, odometry and retrieval. Across transfer and joint-training experiments, supervision that benefits one target can reduce performance on others, with supervised sources transferring within kinematic estimation while, under frozen probing, none approaches the pretrained encoder on language retrieval, and joint-training gains that come at a cost to the paired task. These findings distinguish broad downstream usefulness from complementary task supervision. Code, task cards and a leaderboard are publicly available.\par}
\vskip 1\baselineskip
{\footnotesize\raggedright
\textbf{Code:} \url{https://github.com/SPICExLAB/IMU-Bench}\\
\textbf{Website:} \url{https://spicexlab.github.io/imubench/}\par}
\end{tcolorbox}
\vskip 20pt
\thispagestyle{firststyle}
\begingroup\renewcommand{\thefootnote}{\fnsymbol{footnote}}\footnotetext[1]{Equal contribution.}\endgroup

\section{Introduction}
\label{sec:introduction}
Wrist-worn inertial measurement units (IMUs) provide a common sensing modality for studying human movement. Embedded in everyday smartwatches, accelerometers and gyroscopes capture motion relevant to activity recognition~\citep{yuan2024harnet}, clinical motor assessment~\citep{varghese2024pads} and human--computer interaction~\citep{laput2016viband,wen2016serendipity}, with additional cues for articulated pose estimation and inertial odometry. Like the shadows in Plato's allegory of the cave~\citep[Book VII, 514a--517a]{plato1991republic}, different downstream tasks reveal distinct aspects of the same underlying human motion, each offering only a partial view~\citep{huh2024platonic}. These perspectives raise the question of whether these views share a representation. 

Self-supervised learning has demonstrated substantial reuse of unlabeled wearable recordings. \citet{yuan2024harnet} show transfer from large-scale free-living accelerometry to activity recognition across datasets and populations. Inertia-1 examines objectives, architectures, scale and sensing configurations across activity recognition, freezing-of-gait detection and disease prediction~\citep{xu2026inertia1}. These studies establish transfer across classification tasks and sensing conditions, while leaving less understood how the same representations support articulated joint rotations, metric velocity and motion--language correspondence.

These targets also allow us to examine whether different forms of motion supervision are complementary. Task-transfer studies in vision provide a precedent for measuring such relationships~\citep{taskonomy}. For wrist IMUs, we ask both how broadly pretrained features transfer and whether supervision for one motion target improves or compromises their usefulness for others.

\begin{figure}[!t]
    \centering
    \includegraphics[width=\linewidth]{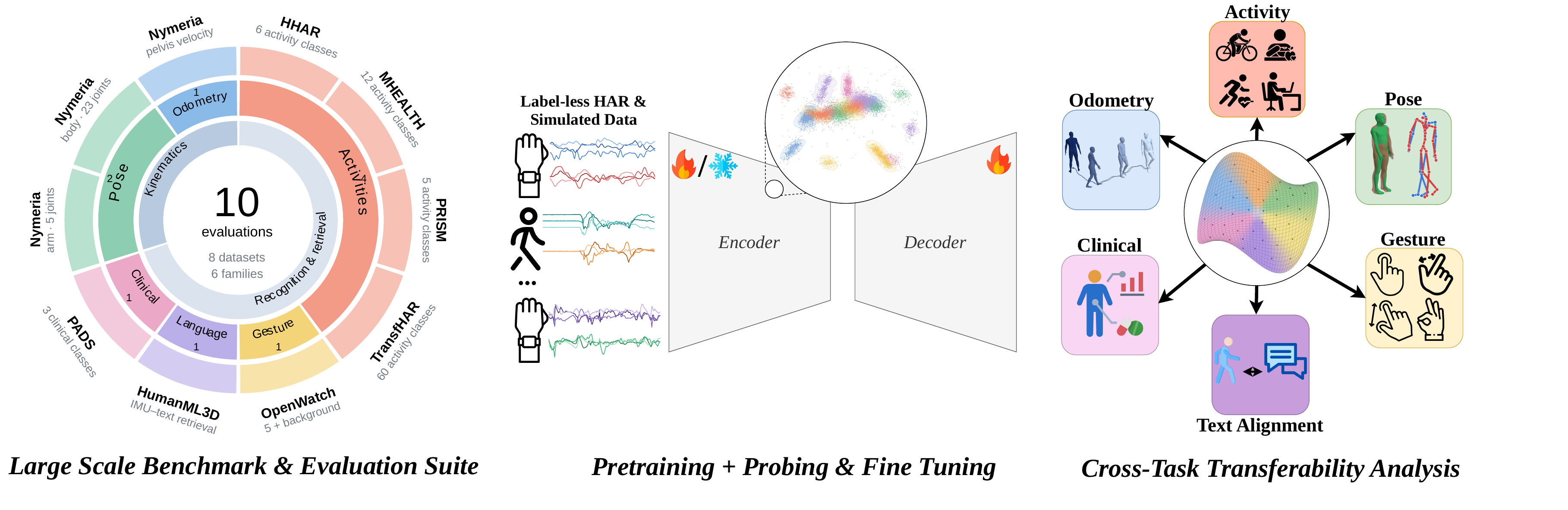}
    \caption{Overview of IMU-Bench. Ten tasks from eight datasets evaluate representations learned from single-wrist accelerometer and gyroscope signals. Pretraining uses unlabeled recorded and simulated motion. Frozen probing and fine-tuning measure downstream performance, while source-task transfer and joint training examine the effects of supervision across tasks. Representation visualizations are schematic.}
    \label{fig:teaser}
\end{figure}

We present IMU-Bench, a benchmark of ten tasks from eight datasets addressing these questions (Figure~\ref{fig:teaser}). We compare six self-supervised methods pretrained on unlabeled activity and daily-living recordings together with simulated wrist motion. Evaluation spans classification, pose and velocity estimation, and IMU--language retrieval, using six-channel inputs from a single wrist. Beyond frozen probing and fine-tuning, we examine supervised source-task transfer and selected joint-training pairs. The results reveal a distinction between broadly useful pretraining and compatible task supervision. Full-body pose and odometry improve over training from scratch after fine-tuning, and each kinematic task's best frozen source is another kinematic task, whereas under frozen probing no supervised source approaches the pretrained encoder on language retrieval. Five of the six kinematic transfers survive, at about half their size, when the tasks are evaluated on disjoint participants, and the observed trade-offs show why individual-task performance alone is insufficient to characterize a shared wrist representation.

We make three contributions:

\begin{itemize}
\item \textbf{A benchmark and comprehensive evaluations.} IMU-Bench brings together ten tasks across six families to evaluate wrist-IMU representations through categorical, kinematic and motion--language prediction targets. We compare six self-supervised methods under matched encoder capacity and common task protocols, alongside training from scratch and retrained task-specific systems.

\item \textbf{A reusable pretrained wrist-IMU model.} We provide InrtlWrist, a masked autoencoder pretrained on unlabeled recorded and simulated wrist motion. Among the six controlled methods, it achieves the best or joint-best frozen-probe performance on nine of ten tasks. Fine-tuning improves on training from scratch on nine of the ten tasks, including full-body pose, odometry and IMU--language retrieval.

\item \textbf{An analysis of cross-task transfer and complementarity.} We measure how source-task supervision and joint training affect other prediction targets. The experiments distinguish beneficial transfer between kinematic tasks from the gap that remains to the pretrained encoder on motion--language retrieval under frozen probing, and identify joint-training gains that come at the expense of the paired task. Additional analyses examine pretraining volume and motion coverage.
\end{itemize}

The benchmark code and the leaderboard are available on the project website, \url{https://spicexlab.github.io/imubench/}; pretrained models will be released there.

\section{Related Work}
\label{sec:related_work}

\paragraph{Wearable sensor benchmarks.}
Wearable-IMU representations are mostly evaluated on activity recognition: pretraining objectives~\citep{haresamudram2022assessing,cai2026benchhar}, cross-dataset generalization~\citep{hong2024crosshar} and foundation models on activity and health outcomes~\citep{yuan2024harnet,narayanswamy2025scaling,abbaspourazad2024largescale,xu2026inertia1,tarale2026bioinspired}; SlotFM adds gesture, gait and sports~\citep{park2025slotfm}, NormWear physiological signals~\citep{luo2026normwear}. Sparse-IMU pose~\citep{dipimu,yi2021transpose,yi2022pip,jiang2022tip,mollyn2023imuposer,mobileposer} and inertial odometry~\citep{chen2018ionet,herath2020ronin,tlio,airio} are studied apart, though label-free physical decoders now link them to self-supervised learning~\citep{leng2026physicalssl}. Task-specific systems for activity~\citep{deepconvlstm}, gestures~\citep{wen2016serendipity,laput2016viband,openwatch}, Parkinson's disease~\citep{bot2016mpower,varghese2024pads} and IMU--language alignment~\citep{unimts,moon2023imu2clip,sensorlm} are our reference points (Appendix~\ref{app:tasks}). None measures how supervision on one task affects the others, which IMU-Bench adds.

\paragraph{Self-supervised learning for wearable sensors.}
Pretext tasks for wearable IMUs include transformation prediction~\citep{saeed2019multitask,yuan2024harnet}, self-training~\citep{tang2021selfhar}, masked reconstruction~\citep{haresamudram2020masked,xu2021limu,he2022mae}, masked modeling with contrast~\citep{simmtm}, factorized multimodal contrast~\citep{focal}, relative contrast~\citep{relcon} and video and text supervision~\citep{das2025primus}; language models can also synthesize virtual IMU~\citep{leng2024imugpt2}. Each came with its own backbone, data and protocol, so we compare five with masked autoencoding under one corpus and matched capacity (Section~\ref{sec:training}). Transformers for multivariate series encode channels independently~\citep{patchtst,moment2024}, as whole-channel tokens~\citep{itransformer} or in separate time and variate stages~\citep{crossformer}; the first loses the accelerometer--gyroscope coupling and the second the patch-level resolution pose and odometry need, hence the alternating frame-wise and global attention of InrtlWrist, adapted from VGGT~\citep{vggt} (Section~\ref{sec:model}).

\paragraph{Task transfer, probing and pretraining data.}
Task relationships have been measured by transferring supervision~\citep{taskonomy,vu2020transferability,task2vec}, by choosing tasks to train together~\citep{caruana1997multitask,standley2020tasks} and by comparing representations directly~\citep{kornblith2019cka,huh2024platonic}; we take the transfer route for wrist IMUs (Section~\ref{sec:results_cross_task}). Because probe accuracy reflects the probe as well as the representation~\citep{alain2016probes,hewitt2019control}, we fix light decoders and add untrained-encoder and raw-input controls (Section~\ref{sec:problem_formulation}). Scaling laws and data pruning~\citep{kaplan2020scaling,sorscher2022pruning} motivated wearable studies of data volume and participant count~\citep{narayanswamy2025scaling,xu2026inertia1}; we vary motion coverage at matched volume (Section~\ref{sec:data_scaling}).

\section{Datasets and Tasks}
\label{sec:datasets_tasks}
\begin{table}[t]
\centering
\small
\caption{Pretraining sources and evaluation datasets. Hours give the size of each released dataset, from which pretraining samples about 280 hours (Section~\ref{sec:pretraining_data}). Parentheses give class counts.}
\label{tab:imubench_tasks}
\vspace{2pt}
\setlength{\tabcolsep}{5pt}
\begin{tabular}{@{}llr@{\hspace{3em}}llr@{}}
\toprule
\multicolumn{3}{@{}l}{\textbf{Pretraining}} & \multicolumn{3}{l@{}}{\textbf{Evaluation}} \\
\cmidrule(r){1-3}\cmidrule(l){4-6}
Regime & Dataset & Hours & Task & Dataset & Hours \\
\midrule
HAR/Everyday & UT-Watch    & 45.3  & Arm pose       & Nymeria   & 300  \\
             & WISDM       & 44.7  & Body pose      & Nymeria   & 300  \\
             & RecoFit     & 43.0  & Odometry       & Nymeria   & 300  \\
             & PAMAP2      & 7.1   & Gesture        & OpenWatch & 1.9  \\
             & SAMoSA      & 6.7   & Retrieval      & HumanML3D & 28.6 \\
             & Shoaib      & 6.5   & Clinical (3)   & PADS      & 35.9 \\
             & OCDetect    & 2,600 & Activity (6)   & HHAR      & 4.8  \\
             & EatingSpeed & 525.9 & Activity (12)  & MHEALTH   & 1.9  \\
Simulated    & Motion-X++  & 40.4  & Activity (5)   & PRISM     & 16.3 \\
             & BONES-SEED  & 144.2 & Activity (60)  & TransfHAR & 2.0  \\
\bottomrule
\end{tabular}
\end{table}
\subsection{Pretraining Data}
\label{sec:pretraining_data}
Pretraining uses 197,094 unlabeled accelerometer--gyroscope windows
(10.24\,s, 50\% overlap) from approximately 280 hours of recorded and
simulated wrist motion. We evaluate transfer to ten tasks from eight
datasets (Table~\ref{tab:imubench_tasks}).
The five pretraining sources are an activity-recognition
pool~\citep{ut_watch,kwapisz2011wisdm,wisdm,recofit,pamap2,samosa,shoaib2016},
OCDetect free-living recordings~\citep{ocdetect}, EatingSpeed
sessions~\citep{eatingspeed,eatradar}, and wrist motion simulated from
Motion-X++~\citep{motionx,motionxpp} and BONES-SEED~\citep{bonesseed}.
Evaluation datasets are excluded from pretraining. Sampling and provenance
checks, including unresolved PRISM--SAMoSA participant overlap, appear in
Appendices~\ref{app:impl} and~\ref{app:audit}.

\subsection{Tasks}
\label{sec:tasks}

\textbf{Pose estimation.}
Nymeria's real right-wrist IMU and motion capture~\citep{nymeria} support
rotation prediction for five right-arm and 23 SMPL joints every 0.32\,s
within 10.24\,s windows. We report arm joint position error (cm) and
root-relative body MPJPE (mm). All three Nymeria tasks share a
recording-session split; participants recur across splits.

\textbf{Inertial odometry.}
Nymeria pelvis velocities predicted in the wrist frame are rotated into
world coordinates using ground-truth initial wrist orientation propagated
by gyroscope integration. Integrated trajectories are scored by absolute
trajectory error (ATE).

\textbf{Microgesture recognition.}
OpenWatch~\citep{openwatch} votes predictions from 1\,s windows into clip
labels, scored by macro-F1. Subject splits use canonical identities after
duplicate and mirrored participant records are merged.

\textbf{IMU--language retrieval.}
Simulated right-wrist HumanML3D clips~\citep{humanml3d} are aligned with CLIP
text embeddings~\citep{clip}. We report mean Recall@1 across both directions
in galleries of 100 on the official test split. Shared source-motion
segments across splits are documented in Appendix~\ref{app:audit}.

\textbf{Clinical classification.}
PADS~\citep{varghese2024pads} assigns 469 participants to healthy control,
Parkinson's disease or differential diagnosis. Inputs use one wrist;
participant predictions average window posteriors across both wrists.
We report participant macro-F1 on a subject-disjoint split.

\textbf{Activity recognition.}
All four activity tasks classify 2.56\,s windows using macro-F1.
HHAR, MHEALTH and PRISM use subject-disjoint
splits~\citep{stisen2015hhar,banos2014mhealth,arakawa2022prism}; PRISM uses its
five global classes. TransfHAR has ten participants, but most activities
were performed by only one~\citep{bradshaw2026transfhar}. Each recording is
therefore split chronologically into 80\% training and 20\% evaluation,
dropping one boundary window.

\section{Methods}
\label{sec:method}

\subsection{Architecture}
\label{sec:model}

Given a wrist-worn window $\mathbf{X}\in\mathbb{R}^{T\times6}$, we divide each accelerometer and gyroscope channel into 0.32\,s patches, forming a $6\times P$ token grid, where $P$ is the number of temporal patches. A 12-layer transformer $f_\theta$ with width $D=384$ and approximately 21 million parameters produces patch representations $\mathbf{H}\in\mathbb{R}^{6P\times D}$.

Following VGGT~\citep{vggt}, layers alternate between frame-wise attention across the six channels at each patch position and global attention across all channels and positions, including a learnable summary token.

\subsection{Training}
\label{sec:training}

We pretrain InrtlWrist by masked reconstruction. The encoder processes visible patches, and a reconstruction decoder predicts the masked measurements. The reconstruction decoder is discarded after pretraining.

To support different input durations, each batch is cropped to 1.28--10.24\,s in multiples of 0.32\,s. Positional embeddings are restricted to the crop, avoiding resizing or interpolation of the position code, following resolution-agnostic vision transformers~\citep{navit}.

We compare against LIMU-BERT (masked reconstruction), SimMTM (reconstruction and contrastive losses), FOCAL (contrastive, orthogonality and temporal losses), RelCon (relative contrastive loss) and SSL-Wearables (transformation-prediction cross-entropy) using a shared pretraining corpus and matched encoder capacity. These objectives use plain attention over the flattened token grid. A plain-attention InrtlWrist variant controls for this architectural difference. Preprocessing and pretraining controls are detailed in Appendix~\ref{app:impl}.

\subsection{Evaluation Protocols}
\label{sec:problem_formulation}

For each task, a decoder $g_\phi$ predicts $\hat{\mathbf{y}}=g_\phi(f_\theta(\mathbf{X}))$. Let $\mathcal{L}_{\mathrm{task}}$ denote the supervised task loss, $\theta_{\mathrm{SSL}}$ pretrained encoder weights and $\theta_{\mathrm{rand}}$ a random initialization. We evaluate frozen probing, fine-tuning and training from scratch under the following optimization settings. Frozen probing fixes the encoder at $\theta_{\mathrm{SSL}}$. Fine-tuning and training from scratch optimize both encoder and decoder, initializing $\theta$ from $\theta_{\mathrm{SSL}}$ and $\theta_{\mathrm{rand}}$, respectively.

\begin{equation}
\underbrace{\min_{\phi}\mathcal{L}_{\mathrm{task}}(\theta_{\mathrm{SSL}},\phi)}_{\text{Frozen probing}} \qquad \underbrace{\min_{\theta,\phi}\mathcal{L}_{\mathrm{task}}(\theta,\phi)}_{\text{Fine-tuning}} \qquad \underbrace{\min_{\theta,\phi}\mathcal{L}_{\mathrm{task}}(\theta,\phi)}_{\text{From scratch}}.
\label{eq:protocols}
\end{equation}

Within each task, controlled comparisons share the decoder architecture, optimizer, training budget, split and model-selection rule. Classification probes use linear heads. Kinematic probes use width-192 attention decoders with two cross-channel and two temporal layers for arm pose, one cross-channel layer for full-body pose, and one temporal layer for odometry. Decoder-capacity sensitivity is examined in Section~\ref{sec:ablations}.

Appendix~\ref{app:tasks} details task definitions, splits and metrics. Each encoder's pretraining corpus is documented; evaluations on datasets included in that corpus are flagged and excluded from matched comparisons. Released external models and retrained task-specific systems are reported separately.
\section{Results}
\label{sec:results}

\begin{figure}[!htb]
\centering
\includegraphics[width=0.85\textwidth]{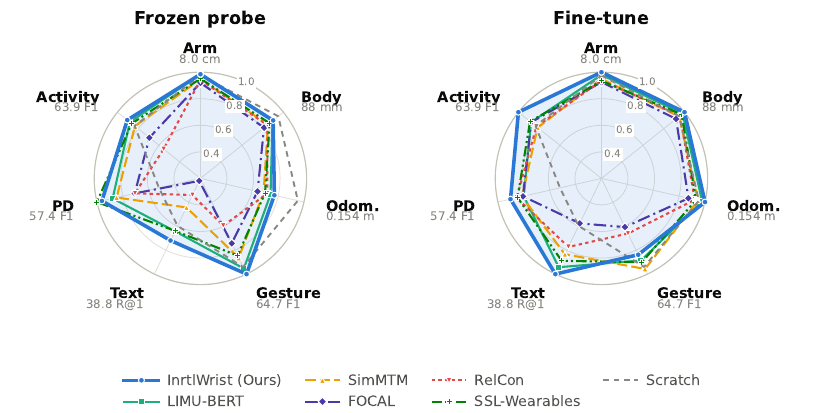}
\caption{\textbf{Downstream performance across task families.} Frozen probing (left) and fine-tuning (right), with training from scratch shown in both panels. Activity averages macro-F1 across four datasets. Each axis is normalized to the best score across both panels, including scratch, using score/best for F1 and Recall@1 and best/error for error metrics. Axis labels give these reference values in the original units. Larger radii indicate better performance; the displayed range is 0.20--1.00.}
\label{fig:main_radar}
\end{figure}

Among the six matched pretraining methods, InrtlWrist achieves the best or joint-best primary score on nine of ten tasks under frozen probing and eight after fine-tuning; fine-tuning improves on training from scratch on nine of ten tasks, gesture being the exception (54.2 against 60.8) (Figure~\ref{fig:main_radar}). Its frozen features outperform training from scratch on mean activity recognition and IMU--language retrieval, but trail it on full-body pose and odometry; fine-tuning surpasses the scratch baseline on both kinematic tasks (Tables~\ref{tab:recognition}--\ref{tab:semantic}).

\subsection{Activity, Gesture, and Clinical Recognition}
\label{sec:results_recognition}

\paragraph{Activity recognition.}
InrtlWrist leads the six matched methods on all four activity datasets under frozen probing, with 57.8 mean macro-F1 against 56.2 for LIMU-BERT and 53.3 for training from scratch (Table~\ref{tab:recognition}). Fine-tuning raises its mean to 63.9, with the largest gains on TransfHAR (72.7 to 86.1) and PRISM (27.9 to 36.8). HHAR instead favors frozen probing, declining from 62.1 to 60.3, while FOCAL achieves the highest fine-tuned score at 61.3. The random frozen encoder reaches 26.2 mean macro-F1, indicating that pretraining improves recognition under the same readout. TransfHAR evaluates later portions of the same recordings; the other three datasets hold out participants.

\begin{table}[!htbp]
\centering
\scriptsize
\setlength{\tabcolsep}{4pt}
\caption{\textbf{Activity, gesture, and clinical recognition.} Macro-F1 and balanced accuracy (BA), reported as percentages with higher values better. Scores are evaluated per window for HAR, per clip for OpenWatch and per participant for PADS. Mean averages the four HAR datasets. Bold marks the best of six matched methods within each protocol, including ties. Gray rows show variants and external references excluded from ranking; $\ddagger$ denotes released NormWear. Task-specific references are retrained DeepConvLSTM for HAR, Mix-Token for OpenWatch and the PADS authors' BOSS pipeline with CatBoost.}
\label{tab:recognition}
\label{tab:main_results}
\begin{tabular*}{\linewidth}{@{\extracolsep{\fill}}lrrrrrrrr@{}}
\toprule
 & \multicolumn{5}{c}{Activity recognition} & Gesture & \multicolumn{2}{c}{PADS} \\
\cmidrule(lr){2-6}\cmidrule(lr){7-7}\cmidrule(lr){8-9}
Encoder / system & HHAR & MHEALTH & PRISM & TransfHAR & Mean & OpenWatch & F1 & BA \\
\midrule
\multicolumn{9}{@{}l}{\emph{Frozen probe}} \\
InrtlWrist (Ours) & \textbf{62.1} & \textbf{68.6} & \textbf{27.9} & \textbf{72.7} & \textbf{57.8} & \textbf{64.7} & 55.2 & 60.0 \\
LIMU-BERT & 61.2 & 67.6 & 25.1 & 71.1 & 56.2 & 61.7 & 50.5 & 58.4 \\
SimMTM & 57.5 & 61.4 & 26.7 & 66.8 & 53.1 & 55.9 & 48.7 & 59.3 \\
FOCAL & 44.9 & 63.0 & 21.9 & 47.0 & 44.2 & 48.1 & 40.7 & 53.9 \\
RelCon & 33.3 & 38.4 & 18.3 & 51.6 & 35.4 & 38.5 & 40.5 & 51.5 \\
SSL-Wearables & 59.2 & 66.3 & 27.6 & 67.4 & 55.1 & 54.7 & \textbf{57.4} & \textbf{61.7} \\
\cmidrule(l){1-9}
\textcolor{gray}{InrtlWrist, plain attn.} & \textcolor{gray}{60.2} & \textcolor{gray}{68.1} & \textcolor{gray}{26.5} & \textcolor{gray}{73.0} & \textcolor{gray}{56.9} & \textcolor{gray}{63.9} & \textcolor{gray}{48.1} & \textcolor{gray}{59.7} \\
\textcolor{gray}{InrtlWrist, accel. only} & \textcolor{gray}{57.8} & \textcolor{gray}{68.5} & \textcolor{gray}{28.0} & \textcolor{gray}{69.5} & \textcolor{gray}{55.9} & \textcolor{gray}{62.2} & \textcolor{gray}{48.2} & \textcolor{gray}{61.6} \\
\cmidrule(l){1-9}
\textcolor{gray}{NormWear$^{\ddagger}$} & \textcolor{gray}{43.7} & \textcolor{gray}{56.1} & \textcolor{gray}{19.2} & \textcolor{gray}{33.7} & \textcolor{gray}{38.2} & \textcolor{gray}{43.8} & \textcolor{gray}{56.6} & \textcolor{gray}{59.7} \\
\cmidrule(l){1-9}
Random frozen encoder & 32.4 & 46.6 & 16.1 & 9.7 & 26.2 & 36.4 & 51.7 & 55.9 \\
Raw input, 0 layers & 5.3 & 1.4 & 15.6 & 0.1 & 5.6 & 11.4 & 12.6 & 33.3 \\
\midrule
\multicolumn{9}{@{}l}{\emph{Fine-tuning}} \\
InrtlWrist (Ours) & 60.3 & \textbf{72.4} & \textbf{36.8} & \textbf{86.1} & \textbf{63.9} & 54.2 & \textbf{51.9} & \textbf{63.9} \\
LIMU-BERT & 56.3 & 62.9 & 29.5 & 75.8 & 56.1 & 57.3 & 46.0 & 55.0 \\
SimMTM & 47.0 & 56.8 & 30.8 & 73.7 & 52.1 & \textbf{61.9} & 46.8 & 59.3 \\
FOCAL & \textbf{61.3} & 65.9 & 28.8 & 68.3 & 56.1 & 39.3 & 46.3 & 57.7 \\
RelCon & 42.9 & 60.3 & 28.9 & 79.0 & 52.8 & 42.7 & 49.1 & 58.1 \\
SSL-Wearables & 53.5 & 64.8 & 32.8 & 75.9 & 56.7 & 58.3 & 48.7 & 59.0 \\
\cmidrule(l){1-9}
\textcolor{gray}{InrtlWrist, plain attn.} & \textcolor{gray}{52.6} & \textcolor{gray}{68.6} & \textcolor{gray}{36.2} & \textcolor{gray}{78.6} & \textcolor{gray}{59.0} & \textcolor{gray}{70.3} & \textcolor{gray}{61.0} & \textcolor{gray}{68.3} \\
\textcolor{gray}{InrtlWrist, accel. only} & \textcolor{gray}{56.8} & \textcolor{gray}{68.3} & \textcolor{gray}{29.8} & \textcolor{gray}{77.2} & \textcolor{gray}{58.0} & \textcolor{gray}{60.6} & \textcolor{gray}{59.9} & \textcolor{gray}{67.7} \\
\cmidrule(l){1-9}
\textcolor{gray}{NormWear$^{\ddagger}$} & \textcolor{gray}{37.1} & \textcolor{gray}{45.6} & \textcolor{gray}{18.3} & \textcolor{gray}{10.3} & \textcolor{gray}{27.8} & \textcolor{gray}{21.0} & \textcolor{gray}{43.1} & \textcolor{gray}{53.3} \\
\midrule
Same backbone (scratch) & 48.8 & 63.2 & 27.8 & 73.5 & 53.3 & 60.8 & 29.9 & 45.5 \\
\midrule
\textcolor{gray}{Task-specific system} & \textcolor{gray}{44.8} & \textcolor{gray}{55.6} & \textcolor{gray}{35.6} & \textcolor{gray}{71.6} & \textcolor{gray}{51.9} & \textcolor{gray}{50.0} & \textcolor{gray}{48.1} & \textcolor{gray}{58.5} \\
\bottomrule
\end{tabular*}
\end{table}

\paragraph{Gesture recognition.}
On OpenWatch, frozen InrtlWrist reaches 64.7 clip macro-F1, compared with 61.7 for LIMU-BERT, 60.8 for scratch and 36.4 for the random encoder. Fine-tuning reduces its score to 54.2, while SimMTM leads the matched methods at 61.9. Frozen InrtlWrist therefore remains ahead of every fine-tuned matched method on this task.

\paragraph{Clinical recognition.}
On PADS, frozen InrtlWrist reaches 55.2 participant macro-F1, below SSL-Wearables at 57.4. Although it exceeds training from scratch by 25.3 points, its advantage over the frozen random encoder is only 3.5 points, providing more limited evidence of a pretraining benefit than on activity and gesture recognition. Fine-tuning reduces macro-F1 to 51.9 while increasing balanced accuracy from 60.0 to 63.9. The retrained BOSS reference reaches 48.1 macro-F1 under the same participant aggregation \citep{varghese2024pads}. All scores use participant-level evaluation; aggregation and class weighting are detailed in Appendix~\ref{app:tasks}.

\subsection{Kinematic Estimation}
\label{sec:results_kinematic}

Frozen InrtlWrist outperforms the random encoder on all three kinematic targets, but remains behind training from scratch on full-body pose and odometry (Table~\ref{tab:kinematics}). Fine-tuning reduces full-body MPJPE from 97.9 to 88.1\,mm and odometry ATE from 0.200 to 0.154\,m, surpassing scratch at 92.7\,mm and 0.162\,m. It achieves the lowest errors among the six matched methods on all five fine-tuned metrics, with LIMU-BERT the closest alternative.

\begin{table}[!htbp]
\centering
\scriptsize
\setlength{\tabcolsep}{2pt}
\caption{\textbf{Kinematic estimation on Nymeria.} Lower values are better for all metrics. Bold marks the best of six matched methods per column, including ties. Spanning cells report each reference under its stated training condition. The gray row reports retrained MobilePoser for pose and Air-IO for odometry, using their own inputs and training recipes. Velocity MSE is reported in units of $10^{-2}$.}
\label{tab:kinematics}
\begin{tabular*}{\linewidth}{@{\extracolsep{\fill}}lrrrrrrrrrr@{}}
\toprule
 & \multicolumn{4}{c}{Arm pose} & \multicolumn{2}{c}{Full-body pose} & \multicolumn{4}{c}{Odometry} \\
\cmidrule(lr){2-5}\cmidrule(lr){6-7}\cmidrule(lr){8-11}
 & \multicolumn{2}{c}{Position (cm)} & \multicolumn{2}{c}{Angle ($^\circ$)}
 & \multicolumn{2}{c}{MPJPE (mm)} & \multicolumn{2}{c}{ATE (m)} & \multicolumn{2}{c}{Vel. MSE ($10^{-2}$)} \\
\cmidrule(lr){2-3}\cmidrule(lr){4-5}\cmidrule(lr){6-7}\cmidrule(lr){8-9}\cmidrule(lr){10-11}
Encoder / system & Probing & Fine-tuning & Probing & Fine-tuning & Probing & Fine-tuning & Probing & Fine-tuning & Probing & Fine-tuning \\
\midrule
InrtlWrist (Ours) & \textbf{8.14} & \textbf{8.01} & 9.89 & \textbf{9.86} & \textbf{97.9} & \textbf{88.1} & \textbf{0.200} & \textbf{0.154} & \textbf{2.732} & \textbf{1.824} \\
LIMU-BERT & \textbf{8.14} & 8.18 & \textbf{9.86} & 9.97 & 98.5 & 89.7 & 0.208 & 0.158 & 2.856 & 1.885 \\
SimMTM & 8.47 & 8.44 & 10.27 & 10.29 & 103.6 & 92.9 & 0.224 & 0.163 & 3.064 & 1.946 \\
FOCAL & 8.71 & 8.66 & 10.70 & 10.69 & 108.4 & 95.9 & 0.240 & 0.177 & 3.334 & 2.218 \\
RelCon & 8.50 & 8.55 & 10.41 & 10.53 & 104.3 & 92.1 & 0.215 & 0.167 & 2.893 & 2.037 \\
SSL-Wearables & 8.42 & 8.57 & 10.14 & 10.58 & 101.7 & 90.6 & 0.220 & 0.166 & 3.078 & 2.030 \\
\midrule
Random encoder (frozen) & \multicolumn{2}{c}{8.46} & \multicolumn{2}{c}{10.35} & \multicolumn{2}{c}{105.0} & \multicolumn{2}{c}{0.233} & \multicolumn{2}{c}{3.217} \\
Same backbone (scratch) & \multicolumn{2}{c}{8.21} & \multicolumn{2}{c}{9.89} & \multicolumn{2}{c}{92.7} & \multicolumn{2}{c}{0.162} & \multicolumn{2}{c}{1.940} \\
\midrule
\textcolor{gray}{Task-specific system} & \multicolumn{2}{c}{\textcolor{gray}{9.74}} & \multicolumn{2}{c}{\textcolor{gray}{24.28}} & \multicolumn{2}{c}{\textcolor{gray}{91.9}} & \multicolumn{2}{c}{\textcolor{gray}{0.180}} & \multicolumn{2}{c}{\textcolor{gray}{3.148}} \\
\bottomrule
\end{tabular*}
\end{table}

Arm pose changes less with fine-tuning. Frozen InrtlWrist and LIMU-BERT tie at 8.14\,cm at the reported precision, compared with 8.21\,cm for scratch. Fine-tuning reduces InrtlWrist's position error to 8.01\,cm, while angular error changes from 9.89$^\circ$ to 9.86$^\circ$, matching frozen LIMU-BERT. The frozen primary-metric ranking therefore does not extend to angular error.

The retrained MobilePoser and Air-IO references reach 91.9\,mm body MPJPE and 0.180\,m ATE, respectively, compared with 88.1\,mm and 0.154\,m for fine-tuned InrtlWrist. Their inputs and training recipes differ from the matched comparison. MobilePoser also receives wrist orientation, and the arm-pose evaluation supplies ancestor joints that it otherwise predicts. These results use session-held-out Nymeria recordings with recurring participants; odometry additionally uses ground-truth initial orientation and short evaluation windows.

\subsection{Semantic Alignment}
\label{sec:results_semantic}

InrtlWrist achieves the highest mean IMU--language Recall@1 among the six matched methods under both protocols (Table~\ref{tab:semantic}). Frozen probing reaches 27.9, compared with 24.9 for LIMU-BERT, 23.2 for training from scratch and 9.5 for the random frozen encoder. Fine-tuning raises the mean to 38.8, with similar improvements in both retrieval directions.

\begin{table}[!htbp]
\centering
\scriptsize
\setlength{\tabcolsep}{3pt}
\caption{\textbf{IMU--language retrieval on HumanML3D.} Recall@1 (\%$\uparrow$), with I and T denoting IMU and text. Mean averages both retrieval directions over 200 sampled galleries of 100 candidates; the full gallery contains 3,701 clips.}
\label{tab:semantic}
\resizebox{\ifdim\width>\textwidth\textwidth\else\width\fi}{!}{%
\begin{tabular}{@{}lrrrrr@{\hspace{8pt}}rrrrr@{}}
\toprule
 & \multicolumn{5}{c}{Frozen probing} & \multicolumn{5}{c}{Fine-tuning} \\
\cmidrule(lr){2-6}\cmidrule(lr){7-11}
 & \multicolumn{3}{c}{Gallery of 100} & \multicolumn{2}{c}{Full gallery}
 & \multicolumn{3}{c}{Gallery of 100} & \multicolumn{2}{c}{Full gallery} \\
\cmidrule(lr){2-4}\cmidrule(lr){5-6}\cmidrule(lr){7-9}\cmidrule(lr){10-11}
Encoder / system & Mean & I$\rightarrow$T & T$\rightarrow$I & I$\rightarrow$T & T$\rightarrow$I
 & Mean & I$\rightarrow$T & T$\rightarrow$I & I$\rightarrow$T & T$\rightarrow$I \\
\midrule
InrtlWrist (Ours) & \textbf{27.9} & \textbf{28.3} & \textbf{27.5} & \textbf{3.6} & \textbf{3.4} & \textbf{38.8} & \textbf{39.5} & \textbf{38.1} & \textbf{5.1} & 4.4 \\
LIMU-BERT & 24.9 & 25.3 & 24.5 & 3.1 & 2.5 & 36.6 & 37.2 & 36.0 & 4.4 & 3.9 \\
SimMTM & 17.2 & 17.4 & 17.0 & 1.7 & 1.2 & 32.4 & 32.7 & 32.1 & 4.0 & 3.4 \\
FOCAL & 8.6 & 9.1 & 8.0 & 0.7 & 0.7 & 22.4 & 22.7 & 22.0 & 2.1 & 1.9 \\
RelCon & 13.2 & 13.9 & 12.4 & 1.0 & 1.0 & 30.0 & 30.5 & 29.5 & 3.2 & 2.9 \\
SSL-Wearables & 24.8 & 25.6 & 24.0 & 2.5 & 2.5 & 34.5 & 34.9 & 34.1 & 4.2 & \textbf{4.5} \\
\midrule
\textcolor{gray}{InrtlWrist, plain attn.} & \textcolor{gray}{26.3} & \textcolor{gray}{26.7} & \textcolor{gray}{26.0} & \textcolor{gray}{3.4} & \textcolor{gray}{2.4} & \textcolor{gray}{37.0} & \textcolor{gray}{37.7} & \textcolor{gray}{36.3} & \textcolor{gray}{4.6} & \textcolor{gray}{4.1} \\
\textcolor{gray}{InrtlWrist, accel. only} & \textcolor{gray}{20.7} & \textcolor{gray}{21.0} & \textcolor{gray}{20.3} & \textcolor{gray}{2.1} & \textcolor{gray}{1.7} & \textcolor{gray}{31.5} & \textcolor{gray}{31.7} & \textcolor{gray}{31.3} & \textcolor{gray}{3.6} & \textcolor{gray}{3.3} \\
\midrule
\textcolor{gray}{NormWear$^{\ddagger}$} & \textcolor{gray}{15.5} & \textcolor{gray}{15.9} & \textcolor{gray}{15.2} & \textcolor{gray}{1.4} & \textcolor{gray}{1.2} & \textcolor{gray}{15.7} & \textcolor{gray}{15.8} & \textcolor{gray}{15.6} & \textcolor{gray}{1.0} & \textcolor{gray}{1.3} \\
\midrule
\multicolumn{11}{@{}l}{\emph{Untrained frozen controls}} \\
Random frozen encoder & 9.5 & 10.0 & 9.1 & 0.6 & 0.4 & -- & -- & -- & -- & -- \\
Raw input, 0 layers & 0.5 & 1.0 & 0.0 & 0.0 & 0.0 & -- & -- & -- & -- & -- \\
\midrule
\multicolumn{11}{@{}l}{\emph{From-scratch references}} \\
Same backbone, scratch & -- & -- & -- & -- & -- & 23.2 & 23.6 & 22.8 & 2.1 & 2.1 \\
\textcolor{gray}{UniMTS (wrist-only)} & -- & -- & -- & -- & -- & \textcolor{gray}{13.2} & \textcolor{gray}{24.9} & \textcolor{gray}{1.5} & \textcolor{gray}{2.3} & \textcolor{gray}{0.0} \\
\bottomrule
\end{tabular}}
\end{table}

The advantage over scratch persists in the full gallery, where fine-tuned InrtlWrist reaches 5.1 IMU-to-text and 4.4 text-to-IMU Recall@1, compared with 2.1 in both directions for scratch. SSL-Wearables achieves the highest text-to-IMU score at 4.5. Although absolute recall falls with the larger candidate set, InrtlWrist remains ahead of scratch in both directions and under both protocols.

The wrist-only UniMTS reimplementation, trained from scratch, reaches 24.9 IMU-to-text Recall@1 but 1.5 in the reverse direction on galleries of 100. This asymmetry is associated with retrieval hubness, detailed in Appendix~\ref{app:tasks}. All encoders receive paired motion--text supervision during downstream training, so these comparisons evaluate the pretrained representation's suitability for learning alignment rather than zero-shot language alignment.
\section{Cross-Task Transferability}
\label{sec:results_cross_task}
\label{sec:transferability}
\label{sec:transfer}

\begin{figure}[!htbp]
    \centering
    \begin{subfigure}[t]{0.495\textwidth}
        \centering
        \includegraphics[width=\linewidth]{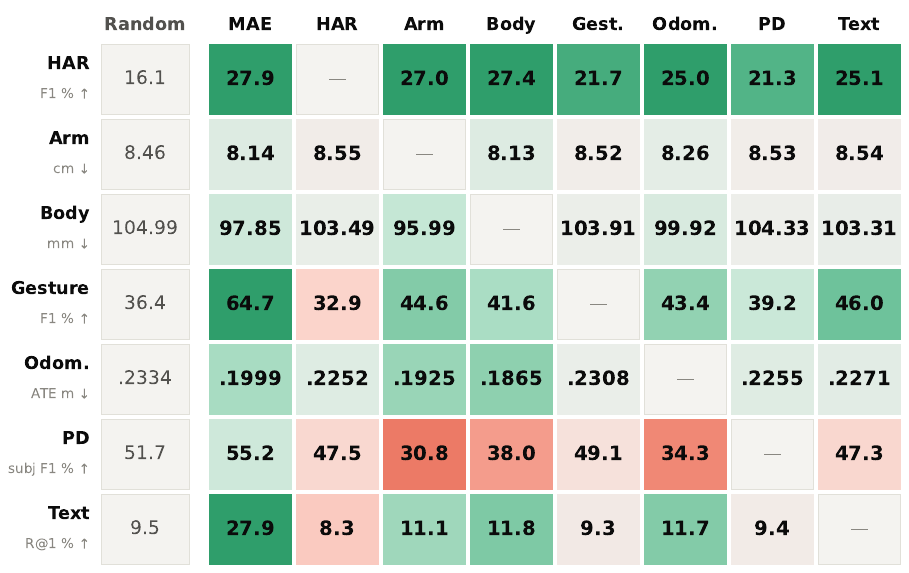}
        \caption{Frozen transfer.}
        \label{fig:transferability_scratch}
    \end{subfigure}\hfill
    \begin{subfigure}[t]{0.495\textwidth}
        \centering
        \includegraphics[width=\linewidth]{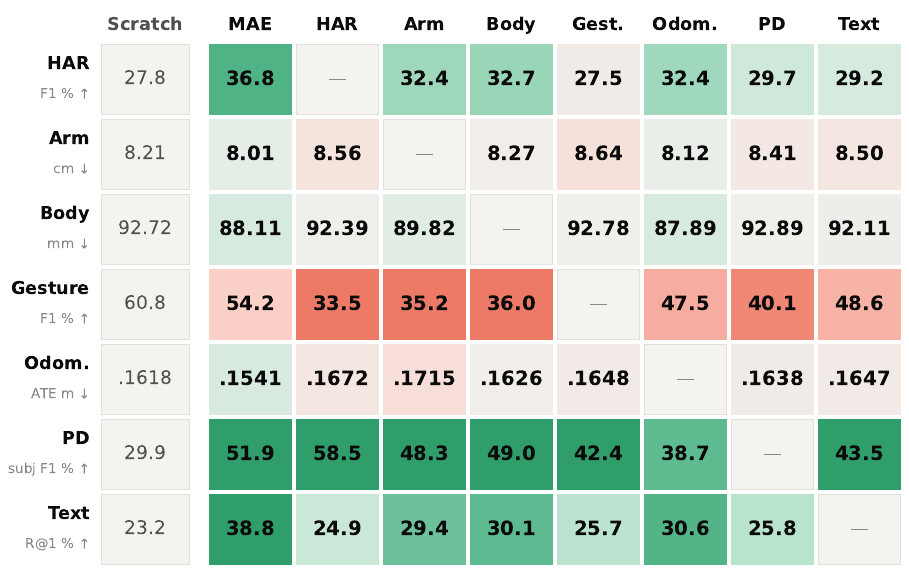}
        \caption{Transfer with fine-tuning.}
        \label{fig:transferability_finetune}
    \end{subfigure}
    \caption{\textbf{Cross-task transfer.} Source encoders (columns) are evaluated on each target (rows). Cells give raw scores; color indicates relative improvement over the first column, using a frozen random encoder in (a) and target training from scratch in (b). Green indicates improvement and orange-pink degradation, with color saturated at $\pm40\%$. MAE denotes InrtlWrist, frozen in (a) and fine-tuned in (b). Dashes omit same-task evaluations. HAR denotes PRISM; PADS uses participant macro-F1.}
    \label{fig:transferability_comparison}
\end{figure}

We evaluate whether supervision on one task produces representations useful for another. Each source encoder is trained from scratch on one of seven tasks, then evaluated on the other six with a new decoder under the benchmark's target splits and budgets (Figure~\ref{fig:transferability_comparison}). Frozen transfer is compared with a random encoder under the same probe, while transfer with fine-tuning is compared with target training from scratch. InrtlWrist, labeled MAE by its pretraining objective, provides the self-supervised reference.

\paragraph{Frozen transfer.}
Supervised source encoders outperform a random frozen encoder in 28 of 42 source–target pairs, including 15 gains above 10\% (Figure~\ref{fig:transferability_scratch}). Kinematic targets benefit most from other kinematic sources: full-body supervision, for example, reduces odometry ATE from 0.233 to 0.186\,m. Because these tasks share Nymeria participants, we repeat the comparison on disjoint participant thirds; five of six kinematic transfers remain positive at roughly half their original magnitude, while MAE leads on full-body pose and odometry (Appendix~\ref{app:participant_split}). All six sources raise PRISM macro-F1 from 16.1 to 21.3--27.4, near MAE's 27.9, but fall below the random encoder on PADS. MAE also leads on gesture and IMU--text (64.7 F1 and 27.9 Recall@1, versus at most 46.0 and 11.8 for supervised sources). Overall, 35 of 42 frozen transfers remain below target training from scratch; the exceptions are full-body-to-arm transfer and the six PADS transfers, for which scratch reaches only 29.9 participant F1.

\paragraph{Transfer with fine-tuning.}
With target fine-tuning, 22 source--target pairs improve over scratch and 20 decline (Figure~\ref{fig:transferability_finetune}). Every source improves IMU--text and PADS, with PRISM supervision reaching 58.5 PADS F1 against 51.9 for direct MAE fine-tuning. Pose and odometry sources improve PRISM by 16--18\% relative to scratch. Four of six sources also improve full-body pose; odometry initialization reaches 87.9\,mm MPJPE, compared with 88.1\,mm for MAE. No source improves odometry, only odometry supervision improves arm pose, and every source reduces gesture F1.

\section{Architecture and Training Ablations}
\label{sec:ablations}
\label{sec:data_scaling}

\begin{figure}[t]
\centering
\begin{minipage}[t]{0.48\linewidth}
\centering
\includegraphics[width=\linewidth]{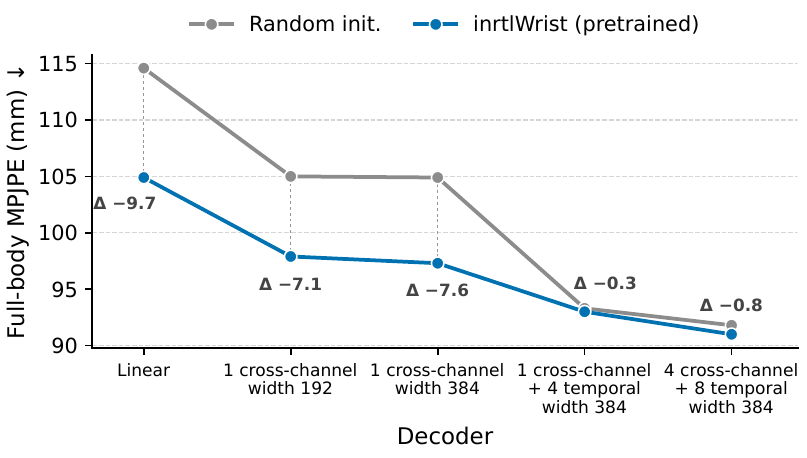}
\caption{\textbf{Decoder capacity.} Frozen-encoder MPJPE (mm$\downarrow$). $\Delta$ denotes pretrained minus random.}
\label{fig:decoder_capacity}
\end{minipage}\hfill
\begin{minipage}[t]{0.48\linewidth}
\centering
\includegraphics[width=\linewidth]{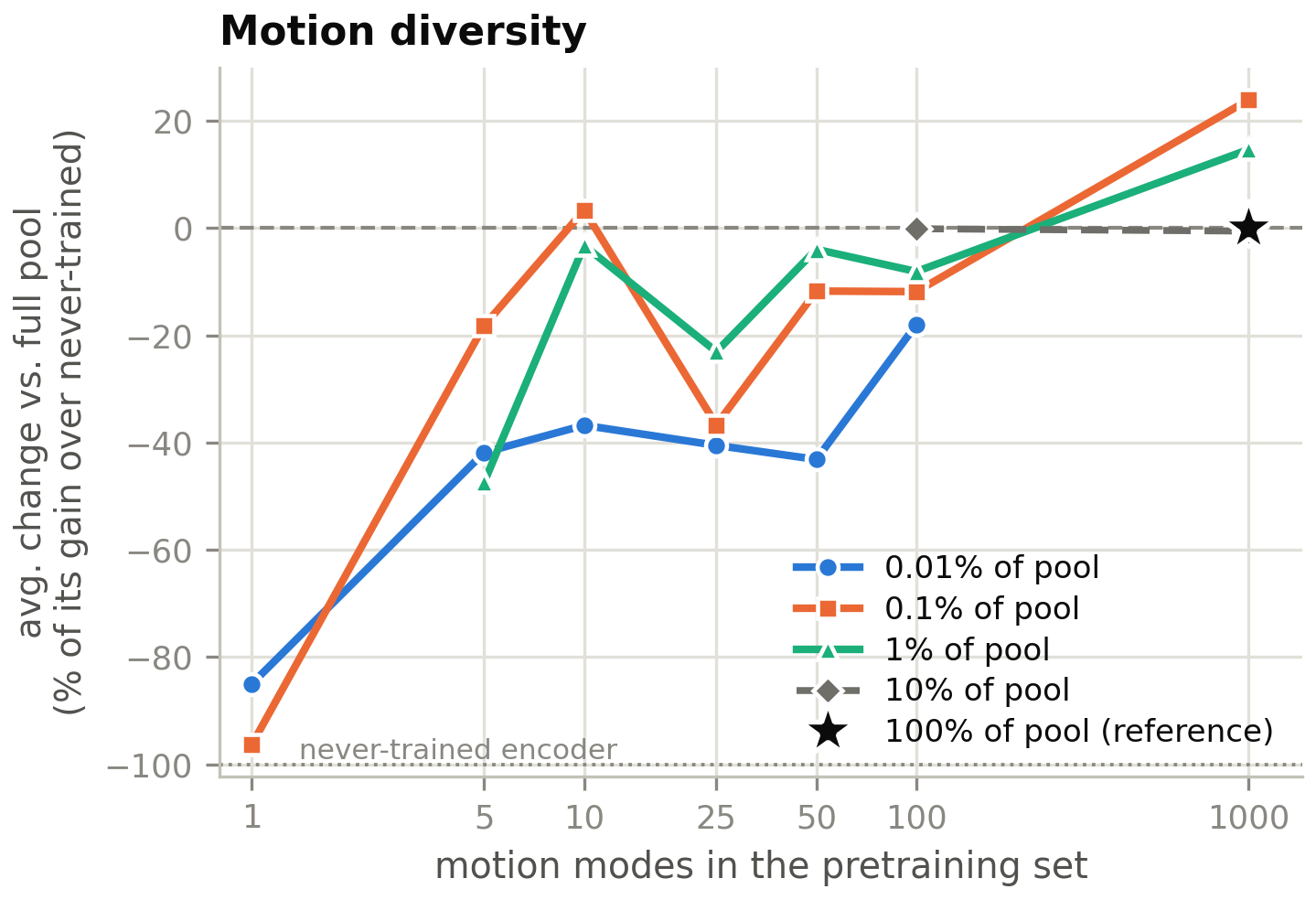}
\caption{\textbf{Motion coverage.} Eight-probe mean after 50,000 updates. Per-probe changes from the full pool are normalized by its gain over random and clipped to $\pm100\%$.}
\label{fig:scaling_avg}
\end{minipage}
\end{figure}

\paragraph{Decoder capacity.} We compare frozen pretrained and random encoders as decoder capacity increases (Figure~\ref{fig:decoder_capacity}). Pretraining reduces body MPJPE by 9.7\,mm with a linear readout and 7.1\,mm with the benchmark decoder. Increasing decoder width changes this gap little, whereas adding temporal layers reduces it below 1\,mm. A 12-layer odometry decoder similarly allows the random encoder to outperform every pretrained encoder (Appendix~\ref{app:controls}). Stronger decoders therefore narrow or reverse the measured advantage of pretraining, motivating the light readouts used in the benchmark.

\paragraph{Pretraining data and motion coverage.} Holding the architecture, probes and training budget fixed, we vary the amount and coverage of pretraining data. We sample uniformly from a five-source pool of 1,959,288 windows or restrict sampling to subsets of 1,000 motion clusters obtained by $k$-means on a separately trained MAE's features. At 50,000 updates, the 0.01\% subsets trail the 0.1\% subsets at every coverage level, while uniform draws from 0.1\% to 100\% of the pool perform alike. Increasing motion coverage generally improves performance, particularly when few clusters are available (Figure~\ref{fig:scaling_avg}). These findings are conditional on the training budget and sampling scheme, since restricting clusters also changes source composition. Appendix~\ref{app:scaling} reports the full grid, normalization and probe protocols.

\paragraph{Joint task supervision.}
We also compare single-task fine-tuning with joint supervision of full-body pose and either odometry or microgesture. Joint training improves odometry ATE from 0.154 to 0.149\,m and gesture F1 from 54.2 to 65.0, but increases body MPJPE and reduces frozen IMU--text Recall@1. Gains on the partner tasks therefore accompany losses on both the jointly trained pose task and language retrieval. Appendix~\ref{app:multitask} reports the schedules and per-task checkpoint selection.

\section{Conclusion}
\label{sec:conclusion}

We present IMU-Bench, ten wrist-IMU tasks evaluated under shared protocols, and the pretrained reference encoder InrtlWrist. Among the six matched methods, InrtlWrist achieves the best or joint-best frozen primary score on nine tasks and outperforms training from scratch on nine after fine-tuning. Source-task supervision transfers between kinematic tasks; on
IMU--text retrieval no frozen supervised source approaches the pretrained encoder. Joint training
improves the partner tasks at a cost to full-body pose and language
retrieval. Decoder capacity can obscure differences between frozen
representations. At the tested pretraining budget, additional broadly
sampled windows provide no consistent downstream gain. These findings
distinguish broad downstream usefulness from mutually beneficial task
supervision.

\paragraph{Limitations.} Our scaling study does not isolate motion coverage from data volume and source composition. It also does not establish how downstream performance
scales with participant count, the balance of recorded and simulated data,
or sensor placement. The benefit of larger corpora at longer training
budgets remains unresolved.

\subsubsection*{Reproducibility statement}
Every number in this paper comes from a registry-driven harness that we release with the benchmark: one task card per task fixing its inputs, decoder, split rule, metric and budget, and one encoder card per model. Appendix~\ref{app:tasks} gives each task's split rule, metric and subject counts, Appendix~\ref{app:impl} the backbone, decoders and optimization settings. Appendix~\ref{app:audit} records the corpus-against-evaluation-set audit, including the duplicated participants and mirrored recordings we had to merge before splitting. Code, task cards and the leaderboard are available on the project page, \url{https://spicexlab.github.io/imubench/}; the per-cell result files and pretrained weights will be released there.

\subsubsection*{Statement on the use of AI}
In this work, we used Anthropic’s Claude through Claude Code to develop and review code, clean and reformat datasets, suggest experimental controls and parameters, analyze results, prepare figures and tables, and assist with the paper’s structure, wording, literature review and references. It also provided feedback on the methodology and helped refine hypotheses. We did not use generative AI to produce synthetic datasets, translate text or conduct qualitative analysis; the simulated wrist IMU was computed from motion capture using a physical model. The authors defined the tasks and protocols, selected the experiments, reviewed the code and results, revised the text, and take responsibility for the final paper.

\FloatBarrier
\bibliography{references}
\bibliographystyle{iclr2027_conference}

\FloatBarrier
\appendix

\section{Implementation Details}
\label{app:impl}

\paragraph{Harness and provenance.} Every number in this paper is produced by one harness from a task card (data, split, input window, decoder, recipe, selection rule, scorer) and an encoder card (checkpoint, architecture, input normalization, pretraining corpus). Each result file records the run, the selected checkpoint step and the scorer, and every table is generated from these files by a script released with the benchmark.

\paragraph{Input.} All tasks use six channels (accelerometer in m/s$^2$, gyroscope in rad/s) from one wrist in a common forward--left--up (FLU) wrist frame. Each window is z-scored per channel before it enters any encoder, in pretraining and downstream alike. During pretraining only, every window is rotated by one of the 24 proper 90$^\circ$-multiple rotations of both sensors~\citep{um2017augmentation}, drawn afresh at each visit; these rotations permute and negate axes, so they commute with per-channel z-scoring. Downstream training uses no augmentation except IMU--text alignment, which applies random 3-D rotations for every encoder alike.

\paragraph{Pretraining.} The self-supervised encoders share one 12-layer, 384-wide ViT over 0.32\,s patches (21\,M parameters), one corpus and one optimizer recipe; they differ in the objective and, for the five external objectives, in using plain rather than alternating frame/global attention.

\paragraph{Controls.} The \emph{random frozen encoder} has the InrtlWrist architecture with random initialization (seed 42) and is never trained. The \emph{raw-input} control is the same model with zero transformer layers, so the decoder reads a frozen random 384-dimensional linear projection of each 16-sample patch plus position embeddings, which preserves the input patch. Both are probed with exactly the decoders, recipes and selection rules of the other rows.

\paragraph{Compute.} Every pretraining and downstream run uses a single NVIDIA A100 or H100 GPU.

\section{Tasks and Metrics}
\label{app:tasks}\label{app:sensor_inputs}

\paragraph{Splits and subjects.} Evaluation is subject-disjoint for microgesture (43 OpenWatch subjects, canonical identities after the duplicated participants of Table~\ref{tab:audit_openwatch} are merged), Parkinson's disease (469 PADS participants) and HHAR, MHEALTH and PRISM (9, 10 and 31 subjects); session-disjoint for the three Nymeria tasks, whose 235 participants recur across sessions; the dataset's own split for IMU--text retrieval; and split in time for TransfHAR (10 subjects), because most of its 60 activities were performed by a single participant (Appendix~\ref{app:userstudy}).

\paragraph{Pose estimation (Nymeria).} Input is the real right-wrist IMU of Nymeria, output one set of 6-D joint rotations per 0.32\,s patch of a 10.24\,s window. The arm task predicts five right-arm joints (collar, shoulder, elbow, wrist, hand), scored by the forward-kinematics position error of these joints with ground-truth ancestors, in cm; the full-body task predicts the 23 SMPL body joints, scored by root-relative MPJPE in mm. Nymeria is split by recording session (867 training and 101 evaluation sessions). All 82 participants of the evaluation sessions also appear in training, in 271 of the 867 training sessions, so the pose and odometry tasks measure generalization to new recordings of seen people; we keep the session split (a participant-disjoint one would remove 31\,\% of the training sessions) and state its scope.

\begin{table}[t]
\centering
\small
\setlength{\tabcolsep}{4pt}
\caption{Arm pose: Nymeria real right-wrist IMU $\rightarrow$ 5-joint right-arm chain rotations. One training seed per cell. Task-specific systems are listed for orientation and are not ranked (Appendix~\ref{app:tasks}). MobilePoser and IMUPoser also read wrist orientation, and must predict the ancestor joints that our arm-pose metric supplies.}
\label{tab:task_arm_pose_nymeria}
\resizebox{\ifdim\width>\textwidth\textwidth\else\width\fi}{!}{%
\begin{tabular}{@{}lrr@{}}
\toprule
Encoder & pos err (cm)$\downarrow$ & ang err ($^\circ$)$\downarrow$ \\
\midrule
\multicolumn{3}{@{}l}{\emph{Frozen probe}} \\
InrtlWrist (Ours) & 8.14 & 9.89 \\
LIMU-BERT & 8.14 & 9.86 \\
SimMTM & 8.47 & 10.27 \\
FOCAL & 8.71 & 10.70 \\
RelCon & 8.50 & 10.41 \\
SSL-Wearables & 8.42 & 10.14 \\
InrtlWrist, plain attention & 8.36 & 10.17 \\
InrtlWrist, accelerometer only & 8.76 & 10.61 \\
\cmidrule(l){1-3}
Random frozen encoder & 8.46 & 10.35 \\
Raw input, 0 layers & 8.51 & 10.30 \\
\midrule
\multicolumn{3}{@{}l}{\emph{Fine-tune}} \\
InrtlWrist (Ours) & 8.01 & 9.86 \\
LIMU-BERT & 8.18 & 9.97 \\
SimMTM & 8.44 & 10.29 \\
FOCAL & 8.66 & 10.69 \\
RelCon & 8.55 & 10.53 \\
SSL-Wearables & 8.57 & 10.58 \\
InrtlWrist, plain attention & 8.19 & 10.06 \\
InrtlWrist, accelerometer only & 8.88 & 10.80 \\
\midrule
Scratch & 8.21 & 9.89 \\
\midrule
\multicolumn{3}{@{}l}{\emph{Task-specific systems} (not ranked)} \\
MobilePoser (Xu et al.) & 9.74 & 24.28 \\
IMUPoser (Mollyn et al.) & 10.15 & 23.82 \\
\bottomrule
\end{tabular}
}
\end{table}

\begin{table}[t]
\centering
\small
\setlength{\tabcolsep}{4pt}
\caption{Full-body pose: Nymeria real right-wrist IMU $\rightarrow$ 23-joint SMPL body (root excluded). One training seed per cell. Task-specific systems are listed for orientation and are not ranked (Appendix~\ref{app:tasks}). MobilePoser and IMUPoser also read wrist orientation.}
\label{tab:task_full_body_pose_nymeria}
\resizebox{\ifdim\width>\textwidth\textwidth\else\width\fi}{!}{%
\begin{tabular}{@{}lr@{}}
\toprule
Encoder & MPJPE (mm)$\downarrow$ \\
\midrule
\multicolumn{2}{@{}l}{\emph{Frozen probe}} \\
InrtlWrist (Ours) & 97.85 \\
LIMU-BERT & 98.55 \\
SimMTM & 103.65 \\
FOCAL & 108.41 \\
RelCon & 104.29 \\
SSL-Wearables & 101.67 \\
InrtlWrist, plain attention & 99.96 \\
InrtlWrist, accelerometer only & 102.79 \\
\cmidrule(l){1-2}
Random frozen encoder & 104.99 \\
Raw input, 0 layers & 106.36 \\
\midrule
\multicolumn{2}{@{}l}{\emph{Fine-tune}} \\
InrtlWrist (Ours) & 88.11 \\
LIMU-BERT & 89.74 \\
SimMTM & 92.91 \\
FOCAL & 95.92 \\
RelCon & 92.15 \\
SSL-Wearables & 90.61 \\
InrtlWrist, plain attention & 88.38 \\
InrtlWrist, accelerometer only & 94.42 \\
\midrule
Scratch & 92.72 \\
\midrule
\multicolumn{2}{@{}l}{\emph{Task-specific systems} (not ranked)} \\
MobilePoser (Xu et al.) & 91.88 \\
IMUPoser (Mollyn et al.) & 96.87 \\
\bottomrule
\end{tabular}
}
\end{table}

\paragraph{Inertial odometry (Nymeria).} The model predicts pelvis velocity in the wrist frame, four values per 0.32\,s patch of a 2.56\,s window. For scoring, each predicted velocity is rotated to the world frame with the ground-truth wrist orientation at the window start propagated by integrating the raw gyroscope, the rotated velocities are summed into a trajectory, and the absolute trajectory error (m) is averaged over all evaluation windows. This uses the true initial orientation and is easier than full inertial navigation; numbers are not comparable to world-frame odometry systems.

\begin{table}[t]
\centering
\small
\setlength{\tabcolsep}{4pt}
\caption{Odometry (wrist frame): Nymeria real right-wrist IMU $\rightarrow$ pelvis velocity in the wrist (FLU) frame. Task-specific systems are listed for orientation and are not ranked (Appendix~\ref{app:tasks}).}
\label{tab:task_odometry_nymeria_local}
\resizebox{\ifdim\width>\textwidth\textwidth\else\width\fi}{!}{%
\begin{tabular}{@{}lrr@{}}
\toprule
Encoder & ATE (m)$\downarrow$ & vel MSE$\downarrow$ \\
\midrule
\multicolumn{3}{@{}l}{\emph{Frozen probe}} \\
InrtlWrist (Ours) & 0.1999 & 0.02732 \\
LIMU-BERT & 0.2078 & 0.02856 \\
SimMTM & 0.2240 & 0.03064 \\
FOCAL & 0.2404 & 0.03334 \\
RelCon & 0.2153 & 0.02893 \\
SSL-Wearables & 0.2205 & 0.03078 \\
InrtlWrist, plain attention & 0.2104 & 0.02892 \\
InrtlWrist, accelerometer only & 0.2296 & 0.03232 \\
\cmidrule(l){1-3}
Random frozen encoder & 0.2334 & 0.03217 \\
Raw input, 0 layers & 0.2558 & 0.03720 \\
\midrule
\multicolumn{3}{@{}l}{\emph{Fine-tune}} \\
InrtlWrist (Ours) & 0.1541 & 0.01824 \\
LIMU-BERT & 0.1576 & 0.01885 \\
SimMTM & 0.1629 & 0.01946 \\
FOCAL & 0.1767 & 0.02218 \\
RelCon & 0.1672 & 0.02037 \\
SSL-Wearables & 0.1656 & 0.02030 \\
InrtlWrist, plain attention & 0.1569 & 0.01880 \\
InrtlWrist, accelerometer only & 0.1898 & 0.02558 \\
\midrule
Scratch & 0.1618 & 0.01940 \\
\midrule
\multicolumn{3}{@{}l}{\emph{Task-specific systems} (not ranked)} \\
TLIO & 0.1874 & 0.02978 \\
Air-IO & 0.1803 & 0.03148 \\
\bottomrule
\end{tabular}
}
\end{table}

\paragraph{Microgesture (OpenWatch).} Five gestures and a background class from 1\,s windows at 100\,Hz, six windows per clip from a centered 2\,s interval, zero-padded to 128 samples. A clip is labeled by the first run of three identical window predictions, otherwise by majority. The subject split is computed over the 43 canonical identities (Appendix~\ref{app:audit}), stratified by posture (33/5/5 identities; 102 validation and 87 test clips); the test split is reported.

\begin{table}[t]
\centering
\small
\setlength{\tabcolsep}{4pt}
\caption{Microgesture recognition on OpenWatch (five gestures plus background, real recordings, held-out test split), our encoder against training from scratch and the task-specific systems.}
\label{tab:task_microgesture_openwatch}
\resizebox{\ifdim\width>\textwidth 0.8\textwidth\else 0.8\width\fi}{!}{%
\begin{tabular}{@{}lrrrr@{}}
\toprule
Encoder & clip F1 (test) (\%)$\uparrow$ & clip acc (test) (\%)$\uparrow$ & win F1 (test) (\%)$\uparrow$ & win acc (test) (\%)$\uparrow$ \\
\midrule
\multicolumn{5}{@{}l}{\emph{Frozen probe}} \\
InrtlWrist (Ours) & 64.7 & 67.8 & 49.4 & 53.6 \\
LIMU-BERT & 61.7 & 60.9 & 49.8 & 51.9 \\
SimMTM & 55.9 & 57.5 & 48.3 & 50.6 \\
FOCAL & 48.1 & 47.1 & 36.9 & 38.3 \\
RelCon & 38.5 & 39.1 & 39.9 & 42.5 \\
SSL-Wearables & 54.7 & 56.3 & 45.6 & 48.9 \\
InrtlWrist, plain attention & 63.9 & 66.7 & 49.4 & 53.3 \\
InrtlWrist, accelerometer only & 62.2 & 64.4 & 49.1 & 51.5 \\
\cmidrule(l){1-5}
Random frozen encoder & 36.4 & 40.2 & 31.9 & 35.4 \\
Raw input, 0 layers & 11.4 & 51.7 & 11.4 & 51.7 \\
\midrule
\multicolumn{5}{@{}l}{\emph{Fine-tune}} \\
InrtlWrist (Ours) & 54.2 & 65.5 & 48.0 & 59.4 \\
LIMU-BERT & 57.3 & 67.8 & 49.1 & 59.4 \\
SimMTM & 61.9 & 69.0 & 50.5 & 59.8 \\
FOCAL & 39.3 & 49.4 & 39.5 & 47.5 \\
RelCon & 42.7 & 57.5 & 38.3 & 51.0 \\
SSL-Wearables & 58.3 & 66.7 & 53.1 & 60.7 \\
InrtlWrist, plain attention & 70.3 & 74.7 & 56.2 & 63.6 \\
InrtlWrist, accelerometer only & 60.6 & 67.8 & 49.7 & 59.2 \\
\midrule
Scratch & 60.8 & 69.0 & 45.1 & 56.5 \\
\midrule
\multicolumn{5}{@{}l}{\emph{Task-specific systems} (not ranked)} \\
Mix-Token (reimpl.) & 50.0 & 62.1 & 49.0 & 59.2 \\
NormWear-Base (first-party) & 45.3 & 51.7 & 42.5 & 50.4 \\
\bottomrule
\end{tabular}
}
\end{table}

\paragraph{IMU--text retrieval (HumanML3D).} Right-wrist IMU is simulated from the HumanML3D motions and aligned with the CLIP ViT-B/32 text space. Recall@1 is the mean of IMU$\rightarrow$text and text$\rightarrow$IMU retrieval over 200 random galleries of 100 test clips with the first caption of each clip, on the 3,701 of the official test split's 4,384 clip ids (2,192 motions and their mirrors, Appendix~\ref{app:audit}) that have simulated wrist IMU and captions in the UniMTS release, carrying 2,721 distinct first captions.

\begin{table*}[t]
\centering
\scriptsize
\setlength{\tabcolsep}{2.5pt}
\caption{IMU--text retrieval on HumanML3D (simulated wrist IMU, test split), where the task-specific row is a wrist-only UniMTS retrained from scratch whose text-to-IMU direction is at chance because of hubness.}
\label{tab:task_language_humanml3d}
\resizebox{\ifdim\width>\textwidth 0.8\textwidth\else 0.8\width\fi}{!}{%
\begin{tabular}{@{}lrrrrrrrrr@{}}
\toprule
 & \multicolumn{1}{c}{} & \multicolumn{4}{c}{Gallery of 100} & \multicolumn{4}{c}{Full gallery} \\
\cmidrule(lr){3-6}\cmidrule(lr){7-10}
Encoder & R@1 mean$\uparrow$ & i2t R@1$\uparrow$ & t2i R@1$\uparrow$ & i2t MRR$\uparrow$ & t2i MRR$\uparrow$ & i2t R@1$\uparrow$ & t2i R@1$\uparrow$ & i2t MRR$\uparrow$ & t2i MRR$\uparrow$ \\
\midrule
\multicolumn{10}{@{}l}{\emph{Frozen probe}} \\
InrtlWrist (Ours) & 27.9 & 28.3 & 27.5 & 42.5 & 41.8 & 3.6 & 3.4 & 9.0 & 8.2 \\
LIMU-BERT & 24.9 & 25.3 & 24.5 & 39.2 & 38.4 & 3.1 & 2.5 & 7.8 & 6.6 \\
SimMTM & 17.2 & 17.4 & 17.0 & 30.3 & 29.8 & 1.7 & 1.2 & 5.0 & 4.2 \\
FOCAL & 8.6 & 9.1 & 8.0 & 19.9 & 18.6 & 0.7 & 0.7 & 2.3 & 2.2 \\
RelCon & 13.2 & 13.9 & 12.4 & 26.9 & 25.0 & 1.0 & 1.0 & 3.6 & 3.1 \\
SSL-Wearables & 24.8 & 25.6 & 24.0 & 39.2 & 37.9 & 2.5 & 2.5 & 7.4 & 6.6 \\
InrtlWrist, plain attention & 26.3 & 26.7 & 26.0 & 40.4 & 39.7 & 3.4 & 2.4 & 8.3 & 7.0 \\
InrtlWrist, accelerometer only & 20.7 & 21.0 & 20.3 & 34.4 & 33.7 & 2.1 & 1.7 & 6.1 & 5.4 \\
\cmidrule(l){1-10}
Random frozen encoder & 9.5 & 10.0 & 9.1 & 21.3 & 19.7 & 0.6 & 0.4 & 2.6 & 2.1 \\
Raw input, 0 layers & 0.5 & 1.0 & 0.0 & 5.2 & 2.0 & 0.0 & 0.0 & 0.3 & 0.1 \\
\midrule
\multicolumn{10}{@{}l}{\emph{Fine-tune}} \\
InrtlWrist (Ours) & 38.8 & 39.5 & 38.1 & 54.9 & 53.9 & 5.1 & 4.4 & 12.6 & 11.0 \\
LIMU-BERT & 36.6 & 37.2 & 36.0 & 52.5 & 51.4 & 4.4 & 3.9 & 11.5 & 10.3 \\
SimMTM & 32.4 & 32.7 & 32.1 & 48.1 & 47.4 & 4.0 & 3.4 & 10.3 & 9.0 \\
FOCAL & 22.4 & 22.7 & 22.0 & 37.2 & 36.5 & 2.1 & 1.9 & 6.2 & 5.7 \\
RelCon & 30.0 & 30.5 & 29.5 & 45.5 & 44.5 & 3.2 & 2.9 & 8.9 & 8.1 \\
SSL-Wearables & 34.5 & 34.9 & 34.1 & 50.4 & 49.5 & 4.2 & 4.5 & 11.1 & 10.5 \\
InrtlWrist, plain attention & 37.0 & 37.7 & 36.3 & 53.0 & 51.8 & 4.6 & 4.1 & 11.7 & 10.6 \\
InrtlWrist, accelerometer only & 31.5 & 31.7 & 31.3 & 46.9 & 46.3 & 3.6 & 3.3 & 9.6 & 8.8 \\
\midrule
Scratch & 23.2 & 23.6 & 22.8 & 38.5 & 37.6 & 2.1 & 2.1 & 6.5 & 6.0 \\
\midrule
\multicolumn{10}{@{}l}{\emph{Task-specific systems} (not ranked)} \\
UniMTS (from scratch, wrist-only) & 13.2 & 24.9 & 1.5 & 39.3 & 6.5 & 2.3 & 0.0 & 6.9 & 0.3 \\
\bottomrule
\end{tabular}
}
\end{table*}

\paragraph{Parkinson's disease (PADS).} Each of the 469 participants is healthy, has Parkinson's disease or a differential diagnosis. PADS records both wrists during eleven scripted assessment steps; each (participant, step, wrist) recording is its own single-IMU session, 22 per participant, resampled to 50\,Hz, so the encoder never sees the two wrists together. Windows of 2.56\,s are classified independently and a participant's prediction is the argmax of the mean posterior over all of that participant's windows; the metric is the macro-F1 over the three classes on a subject-disjoint 80/20 split (375/94 participants; the 94 are 14 healthy, 58 Parkinson's and 22 differential). Every classification head in the benchmark is trained with inverse-frequency class-weighted cross-entropy; it matters most here because a participant's prediction averages about 180 window posteriors, which leaves whatever prior the loss wrote into the head, so an unweighted head assigns all 94 participants to the majority class (25.4 subject macro-F1). The subject metric is the clinically meaningful unit but a coarse one: the never-trained frozen control reaches 51.7, above six of the nine pretrained encoders, so encoders within a few points of each other are not ranked by it. Section~\ref{sec:results} therefore draws its Parkinson's conclusion from the comparison with the dataset authors' pipeline, on which the window and subject units agree. That row is BOSS histograms~\citep{schafer2015boss} with the authors' gradient-boosted classifier (CatBoost~\citep{prokhorenkova2018catboost}), the smartwatch option their paper ranks best~\citep{varghese2024pads} (48.1 subject macro-F1, 58.5 balanced accuracy); it differs from the publication in classifying single-wrist 2.56\,s windows with per-participant averaging rather than one sample per participant on binary tasks.

\begin{table}[t]
\centering
\small
\setlength{\tabcolsep}{4pt}
\caption{Parkinson's classification on PADS (healthy, Parkinson's or differential diagnosis, all values in \%), with the dataset authors' BOSS classifier~\citep{varghese2024pads} re-run under this protocol as the task-specific row.}
\label{tab:task_pd_pads}
\resizebox{\ifdim\width>\textwidth 0.8\textwidth\else 0.8\width\fi}{!}{%
\begin{tabular}{@{}lrrrrr@{}}
\toprule
 & \multicolumn{2}{c}{Per subject} & \multicolumn{3}{c}{Per window} \\
\cmidrule(lr){2-3}\cmidrule(lr){4-6}
Encoder & macro-F1$\uparrow$ & bal-acc$\uparrow$ & bal-acc$\uparrow$ & macro-F1$\uparrow$ & acc$\uparrow$ \\
\midrule
\multicolumn{6}{@{}l}{\emph{Frozen probe}} \\
InrtlWrist (Ours) & 55.2 & 60.0 & 48.3 & 44.1 & 47.9 \\
LIMU-BERT & 50.5 & 58.4 & 49.0 & 43.8 & 46.7 \\
SimMTM & 48.7 & 59.3 & 48.0 & 40.3 & 41.6 \\
FOCAL & 40.7 & 53.9 & 39.2 & 31.3 & 31.5 \\
RelCon & 40.5 & 51.5 & 40.7 & 34.8 & 42.1 \\
SSL-Wearables & 57.4 & 61.7 & 48.5 & 43.9 & 47.3 \\
InrtlWrist, plain attention & 48.1 & 59.7 & 48.8 & 41.4 & 42.6 \\
InrtlWrist, accelerometer only & 48.2 & 61.6 & 47.6 & 40.1 & 41.2 \\
\cmidrule(l){1-6}
Random frozen encoder & 51.7 & 55.9 & 42.7 & 38.4 & 42.3 \\
Raw input, 0 layers & 12.6 & 33.3 & 33.3 & 12.6 & 23.4 \\
\midrule
\multicolumn{6}{@{}l}{\emph{Fine-tune}} \\
InrtlWrist (Ours) & 51.9 & 63.9 & 51.0 & 43.6 & 45.3 \\
LIMU-BERT & 46.0 & 55.0 & 49.3 & 43.0 & 48.9 \\
SimMTM & 46.8 & 59.3 & 49.1 & 40.6 & 41.5 \\
FOCAL & 46.3 & 57.7 & 45.1 & 37.0 & 37.8 \\
RelCon & 49.1 & 58.1 & 46.9 & 39.9 & 41.4 \\
SSL-Wearables & 48.7 & 59.0 & 50.0 & 41.9 & 43.2 \\
InrtlWrist, plain attention & 61.0 & 68.3 & 50.1 & 44.9 & 49.7 \\
InrtlWrist, accelerometer only & 59.9 & 67.7 & 48.7 & 42.6 & 44.7 \\
\midrule
Scratch & 29.9 & 45.5 & 43.9 & 31.8 & 31.9 \\
\midrule
\multicolumn{6}{@{}l}{\emph{Task-specific systems} (not ranked)} \\
PADS pipeline, BOSS + CatBoost (Varghese et al.) & 48.1 & 58.5 & 42.1 & 35.8 & 37.2 \\
\bottomrule
\end{tabular}
}
\end{table}

\paragraph{Activity recognition.} HHAR (6 classes), MHEALTH (12) and PRISM (5 global ADL classes; the 96 fine-grained procedure steps of PrISM-Tracker are a separate, much harder task we do not include) use subject-disjoint splits with 20\,\% of subjects held out, leaving 2, 2 and 6 evaluation subjects (2,661, 1,049 and 15,381 windows); TransfHAR (60 classes) is split in time. All use 2.56\,s windows and report window-level macro-F1 pooled over the evaluation split; Section~\ref{sec:results} also reports the mean over the four sets, and Table~\ref{tab:task_har} gives every metric of every encoder on them.

\begin{table*}[t]
\centering
\scriptsize
\setlength{\tabcolsep}{2.5pt}
\caption{Activity recognition (macro-F1, \%) on the HHAR, MHEALTH and PRISM, which are subject-disjoint, and TransfHAR~\citep{bradshaw2026transfhar}, which is split in time, with their mean.}
\label{tab:task_har}
\resizebox{\ifdim\width>\textwidth\textwidth\else\width\fi}{!}{%
\begin{tabular}{@{}lrrrrrrrrrrrrr@{}}
\toprule
 & \multicolumn{3}{c}{HHAR} & \multicolumn{3}{c}{MHEALTH} & \multicolumn{3}{c}{PRISM} & \multicolumn{3}{c}{TransfHAR} & Mean \\
\cmidrule(lr){2-4}\cmidrule(lr){5-7}\cmidrule(lr){8-10}\cmidrule(lr){11-13}
Encoder & macro-F1$\uparrow$ & bal-acc$\uparrow$ & acc$\uparrow$ & macro-F1$\uparrow$ & bal-acc$\uparrow$ & acc$\uparrow$ & macro-F1$\uparrow$ & bal-acc$\uparrow$ & acc$\uparrow$ & macro-F1$\uparrow$ & bal-acc$\uparrow$ & acc$\uparrow$ & macro-F1$\uparrow$ \\
\midrule
\multicolumn{14}{@{}l}{\emph{Frozen probe}} \\
InrtlWrist (Ours) & 62.1 & 62.7 & 61.9 & 68.6 & 70.0 & 68.3 & 27.9 & 27.2 & 67.0 & 72.7 & 72.2 & 73.0 & 57.8 \\
LIMU-BERT & 61.2 & 60.8 & 59.9 & 67.6 & 68.1 & 66.3 & 25.1 & 25.1 & 66.0 & 71.1 & 70.3 & 71.4 & 56.2 \\
SimMTM & 57.5 & 57.8 & 56.8 & 61.4 & 63.1 & 62.5 & 26.7 & 26.3 & 66.9 & 66.8 & 67.2 & 68.7 & 53.1 \\
FOCAL & 44.9 & 46.7 & 45.8 & 63.0 & 63.0 & 63.4 & 21.9 & 23.2 & 65.9 & 47.0 & 48.1 & 48.9 & 44.2 \\
RelCon & 33.3 & 35.5 & 35.5 & 38.4 & 40.4 & 42.9 & 18.3 & 21.2 & 64.8 & 51.6 & 53.1 & 53.3 & 35.4 \\
SSL-Wearables & 59.2 & 59.7 & 59.5 & 66.3 & 67.0 & 66.3 & 27.6 & 27.1 & 66.3 & 67.4 & 66.6 & 67.2 & 55.1 \\
InrtlWrist, plain attention & 60.2 & 60.4 & 60.3 & 68.1 & 69.6 & 68.5 & 26.5 & 26.1 & 66.4 & 73.0 & 71.4 & 72.7 & 56.9 \\
InrtlWrist, accelerometer only & 57.8 & 58.3 & 58.8 & 68.5 & 70.8 & 69.3 & 28.0 & 27.2 & 66.5 & 69.5 & 69.0 & 70.0 & 55.9 \\
\cmidrule(l){1-14}
Random frozen encoder & 32.4 & 36.2 & 37.7 & 46.6 & 48.9 & 51.8 & 16.1 & 20.2 & 64.1 & 9.7 & 12.1 & 17.7 & 26.2 \\
Raw input, 0 layers & 5.3 & 16.7 & 19.0 & 1.4 & 8.3 & 9.2 & 15.6 & 20.0 & 64.2 & 0.1 & 1.7 & 4.6 & 5.6 \\
\midrule
\multicolumn{14}{@{}l}{\emph{Fine-tune}} \\
InrtlWrist (Ours) & 60.3 & 61.0 & 60.2 & 72.4 & 74.9 & 73.7 & 36.8 & 36.2 & 67.8 & 86.1 & 85.4 & 85.9 & 63.9 \\
LIMU-BERT & 56.3 & 58.5 & 57.2 & 62.9 & 69.2 & 68.1 & 29.5 & 29.2 & 64.9 & 75.8 & 76.3 & 78.3 & 56.1 \\
SimMTM & 47.0 & 48.1 & 48.4 & 56.8 & 62.8 & 62.3 & 30.8 & 30.6 & 67.3 & 73.7 & 74.3 & 74.6 & 52.1 \\
FOCAL & 61.3 & 61.7 & 61.4 & 65.9 & 68.2 & 67.8 & 28.8 & 27.9 & 68.5 & 68.3 & 67.5 & 68.9 & 56.1 \\
RelCon & 42.9 & 46.3 & 46.7 & 60.3 & 63.4 & 62.2 & 28.9 & 28.4 & 68.4 & 79.0 & 78.3 & 79.1 & 52.8 \\
SSL-Wearables & 53.5 & 55.7 & 56.0 & 64.8 & 68.5 & 68.9 & 32.8 & 32.5 & 67.4 & 75.9 & 75.6 & 78.0 & 56.7 \\
InrtlWrist, plain attention & 52.6 & 54.9 & 55.2 & 68.6 & 72.7 & 71.4 & 36.2 & 35.3 & 70.0 & 78.6 & 78.1 & 79.8 & 59.0 \\
InrtlWrist, accelerometer only & 56.8 & 57.9 & 58.0 & 68.3 & 70.7 & 68.8 & 29.8 & 29.5 & 67.5 & 77.2 & 76.6 & 77.8 & 58.0 \\
\midrule
Scratch & 48.8 & 50.3 & 49.6 & 63.2 & 66.2 & 65.2 & 27.8 & 27.2 & 68.5 & 73.5 & 73.4 & 74.4 & 53.3 \\
\midrule
\multicolumn{14}{@{}l}{\emph{Task-specific systems} (not ranked)} \\
DeepConvLSTM (Ordóñez \& Roggen) & 44.8 & 46.6 & 44.8 & 55.6 & 59.5 & 60.2 & 35.6 & 34.0 & 70.6 & 71.6 & 72.9 & 72.8 & 51.9 \\
\bottomrule
\end{tabular}
}
\end{table*}

\paragraph{Task-specific systems.} Published systems (MobilePoser and IMUPoser~\citep{mollyn2023imuposer} for pose, TLIO and Air-IO for odometry, Mix-Token~\citep{openwatch} and the first-party NormWear-Base result~\citep{luo2026normwear} for OpenWatch gestures, UniMTS for IMU--text, DeepConvLSTM~\citep{deepconvlstm} for activity recognition) use their own inputs, output rates, auxiliary components and recipes, so we report them for orientation and do not rank them against the controlled encoders. The pose systems are retrained on the benchmark's Nymeria training sessions and read wrist orientation, which the controlled encoders never see. TLIO and Air-IO are retrained for the wrist-frame odometry protocol and scored by the benchmark's trajectory reconstruction on the same 73,515 evaluation windows. DeepConvLSTM is trained from scratch on the same windows and splits as each activity set's scratch row. The official UniMTS checkpoint is left out because it was pretrained on HumanML3D; the row is its wrist-only reimplementation trained from scratch and scored by the same retrieval script. Its text-to-IMU direction is at chance because of hubness (its text embeddings are nearly collinear, mean pairwise cosine 0.996), not a scoring fault. The secondary metrics of every other cell (angular error, velocity MSE, window and clip accuracy, both retrieval directions at Recall@1, 5 and 10) are in the released results.

\subsection{TransfHAR smartwatch dataset}
\label{app:userstudy}

TransfHAR \citep{bradshaw2026transfhar} contains recordings of ten adult participants (anonymous codes p1--p10) wearing a smartwatch that logged device-motion data at about 50\,Hz: user acceleration and gravity (in $g$, summed and converted to m/s$^2$) and rotation rate (rad/s).
Each of 70 recordings contains one activity (2.0\,h in total; 5,850\,s used for training and 1,284\,s for evaluation). Its file names carry 69 raw activity names; merging spelling variants leaves 60 everyday hand activities (writing, typing, tying a shoelace, stirring coffee, shuffling cards and wiping among them; the dataset's label map lists all 60). Most were performed by a single participant (53 of 60 classes; the other 7 by two to four), so a participant-disjoint split is impossible. The benchmark therefore splits each recording in time, the first 80\,\% for training and the last 20\,\% for evaluation, with one 2.56\,s window removed at the boundary. This measures recognition of an activity later in the same recording, not generalization to new people, and results on this set complement the subject-disjoint sets. The dataset provides the six channels, timestamps, anonymous participant codes and the label map.

\section{Data Audit}
\label{app:audit}

\paragraph{Two kinds of leakage.} \emph{Inductive} leakage places evaluation participants in downstream training and leaks labels; \emph{transductive} exposure shows unlabeled evaluation inputs to self-supervised pretraining and leaks only the test participants' input distribution. Both inflate evaluation numbers.

\paragraph{Pretraining pools.} The reference corpus contains no IMU-Bench evaluation set. Commonly used activity pools, ours included, bundle WISDM~\citep{wisdm}, PAMAP2~\citep{pamap2}, RecoFit~\citep{recofit}, SAMoSA~\citep{samosa}, Shoaib~\citep{shoaib2016}, UT-Watch~\citep{ut_watch} and UTD-MHAD~\citep{utdmhad}, so we exclude these datasets from evaluation; UTD-MHAD is listed in our pool but its clips are shorter than one window and contribute nothing to the corpus, which is why Table~\ref{tab:imubench_tasks} omits it. An earlier eleven-source corpus that we used during development shows how easily exposure happens: it contained the Nymeria and HumanML3D training splits and OpenWatch recordings under the release's random 80/10/10 partition, whose training part contains all five IMU-Bench gesture test participants. That corpus touches this paper only in Appendix~\ref{app:scaling}: in the session-level sweep, reported for reconstruction loss alone; in the encoder whose embeddings define the motion modes; and in the grid's validation set, used only for reconstruction loss and checkpoint selection. The reference encoder is validated on the five pretraining sources alone.

\paragraph{Duplicated participants in OpenWatch.} Grouping the release's files by content hash reveals byte-identical recordings under different participant ids (Table~\ref{tab:audit_openwatch}): 50 ids correspond to 43 distinct identities, and every one of the 35 motion-augmented ids is a wrist-mirrored copy of a real participant, identified by the correlation of acceleration magnitudes. Any split over the release's ids can therefore place the same person in training and test; the original benchmark's split places a copy of training participant 46 in its test set as id 48, and that participant is the best-recognized test participant for every model we trained (clip accuracy 0.88--1.00 against 0.38--0.88 for the other five). With only 8 of 99 test clips, removing it moves clip macro-F1 by just 0.01--0.02, so the score differences in Table~\ref{tab:audit_openwatch} should not be read as the size of the leak. IMU-Bench splits over canonical identities and removes exact duplicates.

\begin{table}[h]
\centering
\small
\caption{OpenWatch release audit. Duplicates are exact: files are grouped by content hash. Bottom: clip macro-F1 (\%) on the release's participant-id split, where test id 48 is a byte-identical copy of training id 46, and on the canonical-identity split used by IMU-Bench. The two splits have different test participants, so the difference mixes leakage with test-set composition (per-participant analysis in the text).}
\label{tab:audit_openwatch}
\begin{tabular}{@{}lr@{}}
\toprule
Participant ids in the release & 50 \\
Distinct identities after merging duplicate ids & 43 \\
Duplicate-identity groups (ids) & {39, 41, 42, 46, 47, 48, 49}; {40, 43} \\
Byte-identical files / groups / ids involved & 254 / 102 / 9 \\
Recordings before / after exact de-duplication & 2,330 / 2,178 \\
Augmented ids that are mirrored copies of real ids & 35 of 35 (ids 51--85; 32 distinct sources) \\
\midrule
Clip macro-F1 (\%) & release split / canonical split \\
\midrule
InrtlWrist, frozen probe & 57.0 / 64.7 \\
InrtlWrist, fine-tune & 64.7 / 54.2 \\
Scratch & 54.8 / 60.8 \\
\bottomrule
\end{tabular}
\end{table}

\paragraph{Motion-capture provenance.} HumanML3D, whose simulated IMU we use for retrieval, is built from 17 AMASS subsets and HumanAct12. The reference corpus contains neither: its Motion-X++ subsets are recovered from video and multi-view capture, and BONES-SEED is studio motion capture (half of the released clips are mirrored; we count only the originals). A direct search for copies, matching every two-second snippet by its bone directions, recovers same-recording HumanML3D clips at a median 4.4$^\circ$ (positive control) but finds no Motion-X++ clip within 5$^\circ$ of any test clip (closest 8.5$^\circ$). BONES-SEED was checked by provenance only; its license requires the attribution that follows: this work uses Motion Data by Bones Studio (\url{https://bones.studio/}); the released encoders' training data includes Motion Data by Bones Studio, and use of the underlying dataset is subject to the BONES Motion Capture Dataset License Agreement. None of the five sources draws on Nymeria. The only overlap we cannot exclude is PRISM with SAMoSA, collected by the same group; identifiers are anonymized per dataset, so shared participants cannot be checked.

\paragraph{Sensor units, and overlap inside HumanML3D.} EatingSpeed stores gyroscope data in deg/s while all other sources use rad/s; per-window z-scoring makes our encoders invariant to this scale, but a pipeline keeping physical units would mix scales by a factor of 57, so the harness converts units on load and rejects a source whose still windows have a median gyroscope magnitude above 0.3\,rad/s or whose 99th percentile exceeds 40\,rad/s. Every HumanML3D split list pairs each clip with a mirrored copy; Table~\ref{tab:imubench_tasks} counts unique motion only. The official split also cuts some AMASS recordings into several clips, so 437 of the 2,192 test clips share frames of a source recording with a training or validation clip; this affects every encoder equally but makes the retrieval task partly transductive.

\section{Controls and Decoder Capacity}
\label{app:controls}

We probed the same frozen encoders on full-body pose with decoders of increasing capacity: a linear readout, the benchmark decoder (one cross-channel attention layer of width 192), one cross-channel layer of width 384, the same plus four temporal layers, and a four-plus-eight-layer decoder of width 384. A never-trained encoder with the InrtlWrist architecture falls 9.7\,mm behind InrtlWrist with the linear readout, 7.1\,mm with the benchmark decoder and 7.6\,mm with the wider single layer, but only 0.3\,mm once four temporal layers are added and 0.8\,mm with the deep decoder; on odometry a 12-layer decoder lets the untrained encoder beat every pretrained one. Deep decoders learn the task from whatever the frozen encoder passes through, so the kinematic probes use the light decoders of the original task configurations (arm pose 2 cross-channel plus 2 temporal layers, full-body pose 1 cross-channel layer, odometry 1 temporal layer, all of width 192). The raw-input control is sensitive to token pooling on recognition and
retrieval; its near-chance scores therefore characterize this readout
configuration rather than learning from raw IMU in general.

\section{Cross-Task Transfer on Disjoint Participants}
\label{app:participant_split}

The three kinematic tasks share one Nymeria split that is session-disjoint but not participant-disjoint (Appendix~\ref{app:tasks}), so a source encoder trained from scratch on one of them (Figure~\ref{fig:transferability_comparison}) has seen the participants of another's evaluation sessions. To separate that exposure from the label supervision we repeated the kinematic block of the figure on three disjoint thirds of the participants, one per task, with every cell (from-scratch source, never-trained control, InrtlWrist probe and transfer probes) retrained inside its third under the benchmark's own decoders, splits and budgets. A third carries a third of the data, so the raw numbers are not comparable with Table~\ref{tab:kinematics}; the gains over the never-trained encoder of the same third are, and Table~\ref{tab:participant_split} lists both. Five of the six frozen transfers survive at about half their size (arm$\rightarrow$full-body $+8.6$ to $+5.2$\,\%, arm$\rightarrow$odometry $+17.5$ to $+10.8$, full-body$\rightarrow$arm $+3.8$ to $+2.3$, full-body$\rightarrow$odometry $+20.1$ to $+11.3$, odometry$\rightarrow$full-body $+4.8$ to $+2.5$); only odometry$\rightarrow$arm changes sign ($+2.4$ to $-2.0$). What does not survive is the comparison with the InrtlWrist: on the shared split a kinematic source matches or beats the frozen InrtlWrist on every kinematic target, whereas on disjoint thirds only full-body$\rightarrow$arm still does (8.62 against 8.68\,cm, inside the run-to-run spread) and the InrtlWrist leads on full-body pose (101.9 against 102.2\,mm) and odometry (0.212 against 0.223\,m); its own margin over the never-trained encoder is $+1.6$, $+5.5$ and $+15.8$\,\% on the thirds against $+3.8$, $+6.8$ and $+14.4$\,\% on the shared split, as expected of a corpus with no Nymeria recording. The sources are also trained on a third of the data, so this control does not separate data volume from participant exposure.

\begin{table}[t]
\centering
\small
\setlength{\tabcolsep}{5pt}
\caption{The kinematic block of Figure~\ref{fig:transferability_comparison} repeated on three disjoint thirds of the Nymeria participants, one per task, so that no encoder has seen a participant that another task is evaluated on. Frozen probes only. \emph{Gain} is the improvement over the never-trained encoder \emph{of the same split}; the raw metrics of the two splits are not comparable, because each third carries a third of the data and every cell, source encoders included, is retrained inside it.}
\label{tab:participant_split}
\begin{tabular}{@{}llrrrr@{}}
\toprule
 & & \multicolumn{2}{c}{session split (Fig.~\ref{fig:transferability_comparison})} & \multicolumn{2}{c}{participant split} \\
\cmidrule(lr){3-4}\cmidrule(lr){5-6}
Target & Frozen encoder & metric & gain & metric & gain \\
\midrule
Arm pose (cm\,$\downarrow$) & never-trained encoder & 8.46 & --- & 8.82 & --- \\
 & InrtlWrist & 8.14 & $+3.8$ & 8.68 & $+1.6$ \\
 & source: full-body pose & 8.13 & $+3.8$ & 8.62 & $+2.3$ \\
 & source: odometry & 8.26 & $+2.4$ & 9.00 & $-2.0$ \\
\midrule
Full-body pose (mm\,$\downarrow$) & never-trained encoder & 105.0 & --- & 107.9 & --- \\
 & InrtlWrist & 97.9 & $+6.8$ & 101.9 & $+5.5$ \\
 & source: arm pose & 96.0 & $+8.6$ & 102.2 & $+5.2$ \\
 & source: odometry & 99.9 & $+4.8$ & 105.1 & $+2.5$ \\
\midrule
Odometry (ATE m\,$\downarrow$) & never-trained encoder & 0.233 & --- & 0.252 & --- \\
 & InrtlWrist & 0.200 & $+14.4$ & 0.212 & $+15.8$ \\
 & source: arm pose & 0.193 & $+17.5$ & 0.224 & $+10.8$ \\
 & source: full-body pose & 0.186 & $+20.1$ & 0.223 & $+11.3$ \\
\bottomrule
\end{tabular}
\end{table}

\section{Joint Task Supervision}
\label{app:multitask}

We vary how the pretrained representation is adapted, asking
whether supervising two tasks at once changes the cross-task trade-offs
observed in Section~\ref{sec:results_cross_task}. Starting from the same MAE, we
alternate updates on full-body pose and either odometry or microgesture,
using separate decoders. Table~\ref{tab:multitask_main} shows gains for
both partner tasks; gesture F1 also exceeds training from scratch.
Full-body MPJPE, however, rises from 88.1\,mm alone to 91.5\,mm with
odometry and 94.4\,mm with microgesture. Frozen IMU--text Recall@1 falls
from 27.9 to 21.0 and 17.2, respectively. These partner-task gains
accompany lower pose and language scores. The task pairs use different training schedules, and each target selects its own checkpoint; the reported scores need not occur at one shared checkpoint. Two pairs were run, each starting from the same reference InrtlWrist as every single-task fine-tune, with each task keeping the benchmark's own loader, decoder and loss and one task per optimizer step in a fixed cycle: full-body pose with odometry at a 1:1 ratio for 40,000 steps, and full-body pose with microgesture at 10:1 for 22,000 steps because the gesture training set is 13 batches.

\begin{table}[!htbp]
    \centering
    \small
    \setlength{\tabcolsep}{5pt}
    \caption{\textbf{Joint fine-tuning with full-body pose.} Both MAE
    conditions start from the same pretrained encoder. Scores use
    per-task checkpoints. Bold: best per row.}
    \label{tab:multitask_main}
    \begin{tabular*}{\linewidth}{@{\extracolsep{\fill}}lrrr@{}}
        \toprule
        Target & Scratch & MAE alone & MAE $+$ pose \\
        \midrule
        Odometry (ATE, m$\downarrow$) & 0.162 & 0.154 & \textbf{0.149} \\
        Microgesture (clip F1$\uparrow$) & 60.8 & 54.2 & \textbf{65.0} \\
        \bottomrule
    \end{tabular*}
\end{table}

\section{Embedding Space of the Reference Encoder}
\label{app:tsne}

\begin{figure}[t]
    \centering
    \includegraphics[width=0.82\linewidth]{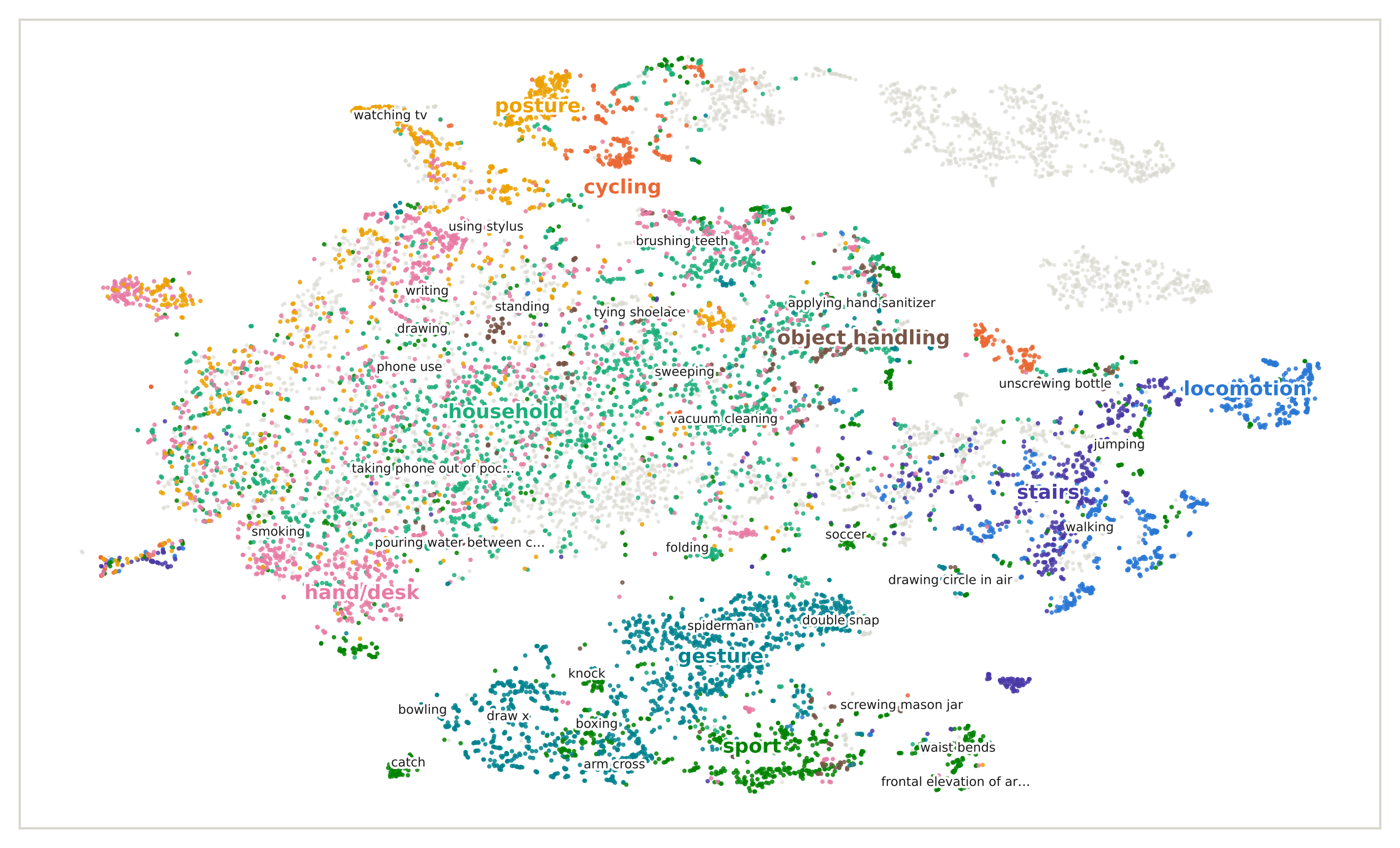}
    \caption{t-SNE~\citep{vandermaaten2008tsne} of the reference InrtlWrist encoder's frozen embeddings (cosine distances): 12,961 windows from eighteen sources, at most 800 each. The 9,139 windows with one of 261 activity labels are colored by semantic family and labeled by family and, where names fit, by activity; gray is the rest, including OCDetect and EatingSpeed, two pretraining sources without activity labels. The projection is illustrative; every statement in the text rests on neighborhood statistics.}
    \label{fig:tsne_fig}
\end{figure}

Figure~\ref{fig:tsne_fig} projects with t-SNE~\citep{vandermaaten2008tsne} the reference encoder's frozen pooled embeddings of 12,961 windows from eighteen sources (twelve labeled activity datasets, the three mocap-derived collections, Nymeria, OCDetect and EatingSpeed). Every number below is the mean fraction of a window's ten nearest neighbors (cosine) that share a property with it, so it measures local geometry, not the projection's layout; the labels are the datasets' own, 261 activities in all.

\paragraph{Activity structure is local and carried by whole-body motion.} 38.8\,\% of a window's ten nearest neighbors share its activity label (chance 1.7\,\%) and 65.6\,\% its semantic family (chance 14.5\,\%), yet neither labeling clusters globally (negative cosine silhouettes), so the projection is not a taxonomy. Restricting the neighborhood to windows from \emph{other} datasets separates what travels between corpora: locomotion 0.512 (chance 0.059), cycling 0.466 (0.026) and stairs 0.415 (0.042), 8.7 to 18 times chance, against household, hand and desk work, posture, gestures and gym exercise at 1.7 to 4.5 times chance. By activity, cycling, running, jogging, walking and stairs travel, while sitting, cooking, typing, washing dishes and folding stay inside the dataset that recorded them, partly because per-window standardization removes the gravity offset that distinguishes static postures.

\paragraph{Synthesis separates the space; pretraining exposure barely does.} The three sources synthesized from motion capture are isolated regardless of exposure (BONES-SEED 0.981 and Motion-X++ 0.899, in the pool; HumanML3D 0.942, held out): mocap-derived windows have 0.96 of their ten nearest neighbors inside their own dataset against 0.59 for recorded windows, a contrast of $+0.364$. The exposure contrast over the recorded windows is $+0.027$ and changes sign under leave-one-source-out (from $-0.060$ without MHEALTH to $+0.090$ without Nymeria); the two least isolated sources are EatingSpeed at 0.335, in the corpus, and Nymeria at 0.338, never seen. Recorded corpora therefore occupy the same region whether or not they were pretrained on, while this objective keeps synthesized inertial data apart even when it trains on a large amount of it, a caution for enlarging a corpus with motion-capture synthesis. UTD-MHAD~\citep{utdmhad} (0.980) and OpenWatch (0.865) are isolated by an artifact: their short clips are wrap-padded 2.91 and 3.41 times to reach 10.24\,s, a periodicity no other source has, so they are excluded from every exposure contrast.

\paragraph{What this supports.} One self-supervised encoder places heterogeneous recorded corpora, seen or unseen, in a shared region with activity structure inside it at about 23 times chance, strongest for gross locomotion and weakest for fine manipulation; it does not support a claim of a dataset-invariant representation, since dataset identity remains the strongest neighborhood signal (0.632 against chance 0.058), and only the task measurements of Sections~\ref{sec:results} and~\ref{sec:results_cross_task} establish what the representation is worth.

\section{Pretraining Volume and Motion Coverage}
\label{app:scaling}

\paragraph{Coverage saturates within the first percent.} A uniform draw of 0.1\% of the pool (1,959 windows) already contains 82\% of the motion modes, a 0.5\% draw 99.7\%, and every draw of 5\% or more all of them; the reference corpus, at 10\% of the pool, covers every mode.

\begin{figure}[t]
    \centering
    \includegraphics[width=0.86\linewidth]{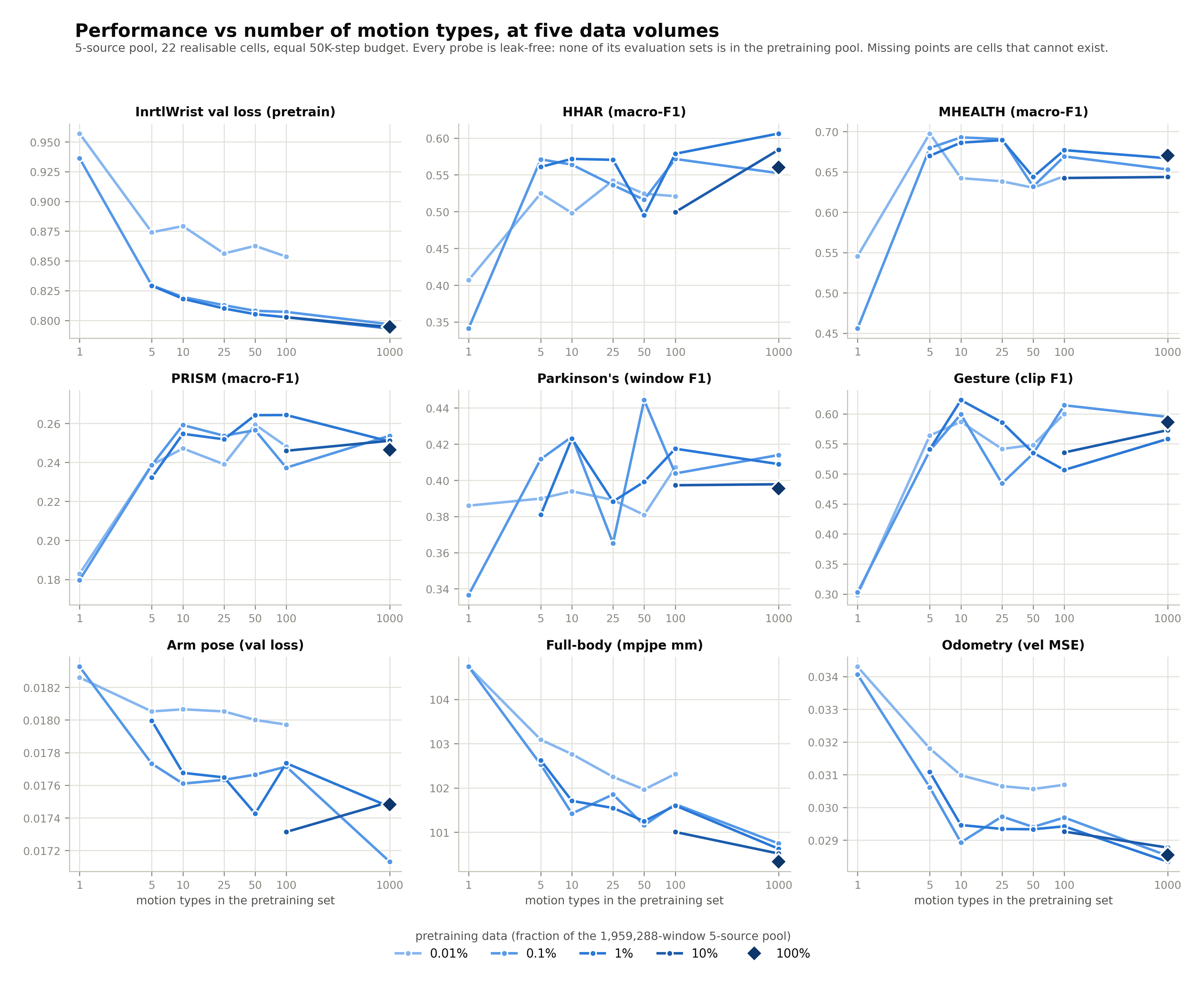}
    \caption{Each probe of the grid against the number of motion modes at five pretraining volumes on the five-source pool, every model trained for 50,000 steps; Figure~\ref{fig:scaling_avg} is its average. Modes are added largest first ($K{=}1$ is 98\,\% EatingSpeed). Activity, Parkinson's (window macro-F1) and gesture panels use the benchmark probes on evaluation sets outside the pool; the arm, full-body and odometry panels use the grid's own lighter probes and recipe, so they show trends within the grid only.}
    \label{fig:grid_clean}
\end{figure}

\paragraph{Downstream performance is flat across a thousand-fold range of data.} With all modes available, pretraining on 1,959 windows gives a reconstruction loss within 0.3\% of the full pool and an arm-pose probe indistinguishable from the other volumes; full-body error varies by 0.4\,mm, odometry velocity error by 1.5\% and MHEALTH, PRISM and gesture F1 by at most 4 points, without a trend. No classification probe improves from 1\% to 100\% of the pool (per task in Figure~\ref{fig:grid_clean}).

\paragraph{Restricting motion modes matters most at the extreme.} At 1,959 windows, training on a single mode raises InrtlWrist loss from 0.797 to 0.936 and takes the eight-probe average to $-96$\% of the full pool's gain, almost back to a never-trained encoder, collapsing every classification probe (gesture F1 30.3 against 59.5 with all modes). From ten modes onward the kinematic probes are within 5\% of the 1,000-mode model in raw error, which for full-body pose and odometry is still 10 to 41\% of the full pool's gain over a never-trained encoder; from five modes onward the activity and gesture probes fluctuate without a trend (Figure~\ref{fig:scaling_avg}). Because modes are added largest first, the mode-count axis also shifts source composition (the single-mode subset is 98\,\% EatingSpeed, the 100-mode subsets about 90\,\% OCDetect), so this comparison bounds the effect of the number of modes rather than isolating it.

\paragraph{Scope and caveats.} These results hold for one architecture and one data pool at 50,000 steps, and they do not separate diversity from volume: the uniform draws change size and coverage together, and the mode-restricted draws change source composition with the mode count. Enlarging the corpus brings no consistent downstream gain at this budget, coverage of the motion modes is complete long before the corpus is, and whether more data pays at larger budgets is open. The motion modes are $k$-means clusters of embeddings from an InrtlWrist trained on the earlier eleven-source corpus, re-clustered on the five-source pool: a data-driven partition, not a taxonomy of movements. An earlier session-level sweep on the eleven-source corpus (5\,\%, 25\,\% and 100\,\% of 3.56 million windows, 300,000 steps each) lowers reconstruction loss by only 0.4\,\% (0.7604 to 0.7572); because that corpus contains evaluation participants (Appendix~\ref{app:audit}), we do not use its downstream results.

\end{document}